\documentclass[runningheads]{llncs}

\usepackage{eccv}

\usepackage{eccvabbrv}

\usepackage{graphicx}
\usepackage{colortbl}
\usepackage{booktabs}
\usepackage{multirow}
\usepackage{multicol}
\usepackage{tabularray}
\usepackage{arydshln}
\usepackage{booktabs}       %
\usepackage{amsfonts}       %
\usepackage{nicefrac}       %
\usepackage{microtype}      %
\usepackage{xcolor}         %
\usepackage{amsmath,bm}
\usepackage{mathtools}
\usepackage{pifont}
\usepackage{dashrule}
\usepackage[accsupp]{axessibility}  %
\usepackage{comment}

\newcommand{\name}{{RICA$^2$}\xspace}

\newcommand{\namedet}{{RICA$^2\dagger$}\xspace}

\newcommand{\squishlist}{
	\begin{list}{$\bullet$}
		{ \setlength{\itemsep}{0pt}
			\setlength{\parsep}{1pt}
			\setlength{\topsep}{1pt}
			\setlength{\partopsep}{0pt}
			\setlength{\leftmargin}{1.5em}
			\setlength{\labelwidth}{1em}
			\setlength{\labelsep}{0.5em} } }
\newcommand{\squishend}{\end{list} 
}

\usepackage{hyperref}

\usepackage{orcidlink}

\begin{document}

\title{RICA$^2$: Rubric-Informed, Calibrated\\ Assessment of Actions}

\author{Abrar Majeedi\orcidlink{0000-0002-5615-5613} \and
Viswanatha Reddy Gajjala\orcidlink{0000-0001-7256-9659} \and
 Satya Sai Srinath Namburi GNVV\orcidlink{0009-0002-0726-5653} \and Yin Li\orcidlink{0000-0003-4173-9453}}

\authorrunning{A. Majeedi et al.}
\institute{University of Wisconsin-Madison, Madison Wisconsin 53706 USA \email{\{majeedi,vgajjala,sgnamburi,yin.li\}@wisc.edu}}

\maketitle
\begin{abstract}
The ability to quantify how well an action is carried out, also known as action quality assessment (AQA), has attracted recent interest in the vision community. Unfortunately, prior methods often ignore the score rubric used by human experts and fall short of quantifying the uncertainty of the model prediction. To bridge the gap, we present \name{} --- a deep probabilistic model that integrates score rubric and accounts for prediction uncertainty for AQA. Central to our method lies in stochastic embeddings of action steps, defined on a graph structure that encodes the score rubric. The embeddings spread probabilistic density in the latent space and allow our method to represent model uncertainty. The graph encodes the scoring criteria, based on which the quality scores can be decoded. We demonstrate that our method establishes new state of the art on public benchmarks, including FineDiving, MTL-AQA, and JIGSAWS, with superior performance in \textit{score prediction} and \textit{uncertainty calibration}. Our code is available at~\url{https://abrarmajeedi.github.io/rica2_aqa/}.

\end{abstract}
    
\section{Introduction}
\label{sec:intro}

Action quality assessment (AQA), aiming at quantifying how well an action is carried out, has been widely studied across scientific disciplines due to its broad range of applications. AQA is key to sports science and analytics. The right way of performing actions maximizes an athlete's performance and minimizes injury risk. AQA is crucial to occupational safety and health. High-quality actions mitigate the physical stress and strain in the workspace. AQA is pivotal for physical therapies. The quality of actions reveals the progress in rehabilitation. AQA also plays a major role in surgical education. Proficient actions improve the outcome and reduce complications. 

Observational methods for AQA have been well established for various tasks, \eg, gymnastics~\cite{prassas2006biomechanical}, manual material handling~\cite{waters1994applications}, and surgery~\cite{martin1997objective}. These methods involve a human expert observing an action and decomposing it into a series of key steps. Each of these steps, or a subset of them, can be grouped into a factor and then evaluated using a Likert scale~\cite{likert1932technique} following a pre-defined criterion. Ratings for individual factors, sometimes complemented with impression-based global ratings, are then summarized into a final quality score~\cite{martin1997objective}. 
Multiple expert ratings are often considered to account for the variance in the scores. While these methods are commonly adopted, they require significant input from human experts and are thus costly and inefficient.

\begin{figure}[t]
    \centering
    \includegraphics[width=0.98\columnwidth]{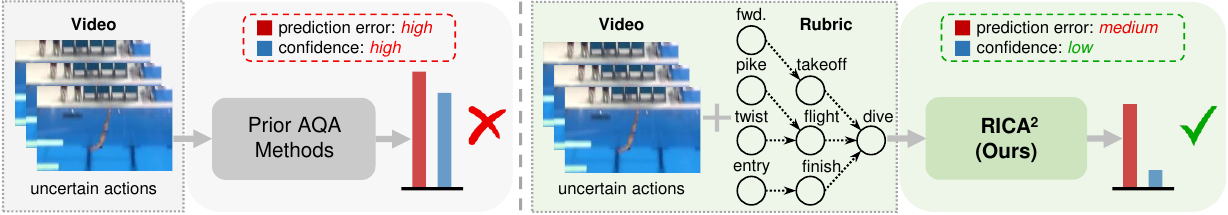}
    \caption{
    \name integrates score rubric used by human experts and accounts for prediction uncertainty, resulting in \textit{accurate} predictions and \textit{calibrated} uncertainty estimates. 
    }
    \label{fig:teaser}
\end{figure}

There is a burgeoning interest in the vision community to develop video-based AQA~\cite{tang2020uncertainty, pirsiavash2014assessing, parmar2017learning, xu2019learning}. While current solutions have made steady progress across benchmarks~\cite{xu2022finediving, parmar2019and, gao2014jhu}, their decision-making processes differ largely from prior observational methods. Almost all prior solutions learn deep models to directly map input videos to scores. Many of them employ an exemplar-based approach, in which a model predicts relative scores by referencing exemplar videos with similar actions and known scores~\cite{bai2022action, xu_core_2021, xu2022finediving}. Few of them have considered the structure of the actions or their scoring criteria used by observational methods. 

Further, existing AQA methods face a key challenge in the accurate quantification of model uncertainty \ie, the uncertainty of model prediction that is calibrate to the expected error~\cite{guo2017calibration}. Knowing this uncertainty is particularly helpful for AQA, \eg, when assessing the quality of high-stakes competitions or surgical procedures. With proper calibration, 
videos that have uncertain predictions can be passed to human experts for a thorough evaluation. Several recent works have started to consider the variance among scores from multiple human experts~\cite{tang2020uncertainty, zhou2022uncertainty, zhang2023auto, zhouHeirarchicalGCNaqa}. Unfortunately, they still fall short of considering prediction uncertainty, leaving this challenge largely unaddressed. 

To bridge the gap, we develop a deep probabilistic model for AQA by integrating score rubrics and modeling the uncertainty of the prediction (see~\cref{fig:teaser}). Central to our method lies in the stochastic embedding of action steps, defined on a graph structure that encodes the score rubric. The embeddings spread probabilistic density in latent space and allow our method to represent model uncertainty. The graph encodes the scoring criteria, based on which the quality scores can be decoded. We also present a training scheme and describe an approach to estimate uncertainty. Putting things together, our method, dubbed \name (Rubric-informed, Calibrated Assessment of Actions), yields accurate action scores with additional uncertainty estimates.

We evaluate \name on several public AQA datasets, covering sports and surgical videos. 
Particularly, \name establishes new state of the art on FineDiving~\cite{xu2022finediving}, MTL-AQA~\cite{parmar2019and} and JIGSAWS~\cite{gao2014jhu}. On FineDiving~\cite{xu2022finediving} -- the largest and most challenging AQA benchmark, \name outperforms latest methods in prediction accuracy (a boost of +0.94\% in Spearman's Rank Correlation Coefficient ($SRCC$)) and demonstrates significantly improved uncertainty calibration (a gain of $+0.178$ in Kendall Tau~\cite{kendalltau}). Similarly, on MTL-AQA~\cite{parmar2019and}, the most commonly used dataset for AQA, \name attains state-of-the-art $SRCC$, and again largely improved calibration (a gain of $+0.444$ in Kendall Tau). On JIGSAWS~\cite{gao2014jhu}, \name beats the previous best results by a relative margin of $+3.37\%$ in $SRCC$. Further, we present extensive experiments to evaluate the key design of \name. 

\smallskip
Our main \textbf{contributions} are summarized into three folds.
\squishlist
    \item We propose \textit{RICA$^2$}, a novel deep probabilistic method that incorporates scoring rubrics and uncertainty modeling for AQA, resulting in accurate scores and calibrated uncertainty estimates.
    \item Our technical innovations lie in (a) a graph neural network to model the scoring rubric in conjunction with stochastic embeddings on the graph to account for prediction uncertainty and (b) a training scheme under the variational information bottleneck framework.
    \item Our extensive set of experiments demonstrates that \name achieves state-of-the-art results in AQA, significantly outperforming prior methods in both prediction accuracy and calibration of uncertainty estimates. 
\squishend

\section{Related Work}
\label{sec:related_work}

\textbf{Action quality assessment (AQA).} Early works in AQA~\cite{gordon1995automated, pirsiavash2014assessing} employed handcrafted features to estimate quality scores in videos. More recent methods developed various deep models, including convolutional~\cite{tang2020uncertainty,xu_core_2021}, graph~\cite{pan2019action}, recurrent~\cite{parmar2017learning, xu2019learning}, and Transformer~\cite{bai2022action,xu2022finediving} networks. AQA has also been widely considered in surgical education~\cite{liu2021towards}, rehabilitation~\cite{qiu2022pose}, and ergonomics~\cite{chen2013automated}.

Recently, exemplar-based methods~\cite{bai2022action, xu_core_2021, xu2022finediving} have emerged as a promising solution for AQA due to their impressive performance across benchmarks. These methods predict the relative score of an input video by comparing it to selected exemplar videos with similar action steps and known scores. A limitation of this paradigm is the requirement of exemplar videos at inference time. This strategy largely deviates from existing observational methods used by human experts and leads to significantly higher computational costs. While \name also uses action steps in the input video, it further integrates the scoring rubric of these steps and offers a solution for no-reference AQA \ie \textit{without using exemplars}.

Several recent works have started to consider the modeling of score uncertainty in AQA~\cite{tang2020uncertainty,zhou2022uncertainty, zhang2023auto, zhouHeirarchicalGCNaqa}. For example, Tang et al.\ \cite{tang2020uncertainty} proposed to model the final scores using a Gaussian distribution. They presented a model (MUSDL) trained to predict the score distribution. This distribution learning idea was further extended in~\cite{zhou2022uncertainty, zhang2023auto, zhouHeirarchicalGCNaqa}. However, modeling the score distribution does not warrant the quantification of model uncertainty, as the output distributions might not be calibrated with prediction errors. While \name also predicts a Gaussian distribution for the scores, our key design is to consider stochastic embeddings to quantify prediction uncertainty, resulting in \textit{calibrated uncertainty estimates}.

The most relevant work is IRIS~\cite{matsuyama2023iris}. IRIS incorporates score rubric into a convolutional network for AQA. This is done by segmenting key steps in the video and predicting sub-scores for individual steps. Similar to IRIS, \name also considers rubric in a deep model. However, \name adapts a graph network, treats sub-scores as latent embeddings, predicts the final score, and further quantifies prediction uncertainty. These differences allow \name to be trained on major public datasets with only final scores, and to output calibrated uncertainty estimates, both of which cannot be achieved by IRIS.

\smallskip
\noindent \textbf{Modeling uncertainty with stochastic embedding.} Stochastic embedding, initially introduced in NLP~\cite{neelakantanSPM14,vilnis2014word}, treats each embedding as a distribution. This approach has gained recent attention for modeling uncertainty in deep models. Oh et al.\ \cite{oh2018modeling} considered probabilistic embeddings for metric learning and proposed to model uncertainty based on the stochasticity of embeddings. This idea was further adopted in many vision tasks, including face verification~\cite{shi2019probabilistic}, age estimation~\cite{li2021learning}, pose estimation~\cite{sun2020view}, and cross-modal retrieval~\cite{chun2021probabilistic}. Another related line of work is the conditional variational autoencoder~\cite{sohn2015learning}, where a probabilistic representation of the input is used for a prediction task. 
Our approach shares a similar idea of using stochastic embeddings to model uncertainty yet is specifically designed for AQA. Our method significantly extends prior idea to embed action steps on a graph structure, and to propagate these stochastic embeddings on the graph. 

\smallskip
\noindent \textbf{Graph neural networks (GNNs).} 
GNNs~\cite{scarselli2008graph,duvenaud2015convolutional,kipf2017semisupervised} offer a powerful tool to leverage the relational inductive bias inherent in data~\cite{battaglia2018relational,Xu2020What}. This inductive bias is beneficial to aggregate a global representation from a group of local ones~\cite{santoro2017simple}. Recently, Zhou et al.\ \cite{zhouHeirarchicalGCNaqa} proposed a hierarchical graph convolutional network for AQA, in which a GNN was used for video representation learning. In contrast, we adapt graph networks to model score rubrics used by observational methods. 

\section{AQA with Score Rubric and Uncertainty Modeling}

Our goal is to assess the quality of an action within an input video. Let $X$ be the video with the action and $Y$ as its quality score. Our method further considers the structure of the action and a scoring rubric based on the structure. 

\smallskip
\noindent \textbf{Action steps}. We assume that the action in $X$ comprises a known, ordered set of key steps, denoted as $\mathbb{S}=(s_1, s_2, ..., s_k)$. Each $s$\footnote{For the sake of brevity, we omit the subscript as long as there is no confusion.} represents a necessary sub-action for successfully executing the action. Further, $s$ is associated with a text description that elucidates the specifics of the corresponding step, \eg, ``a front-facing takeoff'' for diving. This assumption is especially well suited for structured actions, such as diving or surgery, where the key steps are predetermined and follow a specific sequence. Note that the timing of the key steps is not presumed. Even if key steps are unavailable, they can be detected using action recognition methods~\cite{carreira2017quo,feichtenhofer2019slowfast} (see supplement \textcolor{blue}{Sec.\ C.4}).

\begin{figure*}[t]
    \centering
    \includegraphics[width=0.95\linewidth]{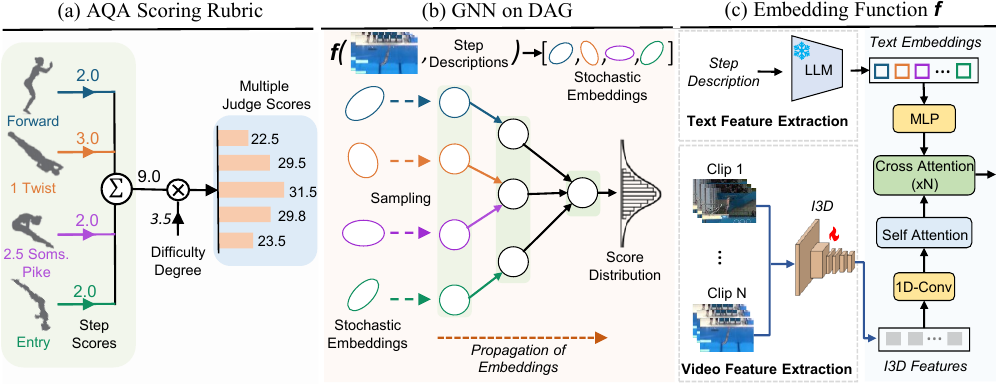}
     \caption{\textbf{Overview of \name.} Leveraging scoring rubrics (a), \name integrates a graph representation of action step and rubric with uncertainty modeling (b). Specifically, \name takes an input of the video and its key action steps, encodes the input into embeddings (c), refines the embeddings through a deep probabilistic model, and outputs an action score in tandem with its uncertainty estimate.
    } %
    \label{fig:architecture}
\end{figure*}

\smallskip
\noindent \textbf{Scoring rubric}. We further assume a pre-specified scoring rubric based on the key steps --- a common strategy in technical skill assessment~\cite{prassas2006biomechanical,waters1994applications,martin1997objective}. Specifically, each action step $s_k$ is independently scored, \ie, $s_k \mapsto y_k$. Subsequently, a rule-based rubric is employed to aggregate individual scores $\{y_k\}$ and calculate a final quality score $Y$, \ie, $\{y_k\}\mapsto Y$, in which steps might be grouped into intermediate stages (see an example in~\cref{fig:architecture} (a-b)). This rubric follows a deterministic yet often non-injective mapping, \eg a many-to-one mapping such as summation. 

\smallskip
\noindent \textbf{Method overview}.
We now present \name --- a deep probabilistic model for AQA that leverages known action steps and incorporates the scoring rubric for modeling. Importantly, \name accounts for prediction uncertainty, \ie, when the model prediction \textit{can} and \textit{cannot} be trusted. \cref{fig:architecture} presents an overview of \name. It consists of two main model components: (a) a graph neural network that integrates the key steps and scoring rubric (\cref{sec:graph}); and (b) stochastic embeddings defined on the graph to capture prediction uncertainty (\cref{sec:stochastic}), coupled with (c) a learning scheme under the variational information bottleneck framework (\cref{sec:vib}). In what follows, we delve into the details of \name. 

\subsection{Integrating Actions Steps and Scoring Rubric with Graph}\label{sec:graph}
\noindent \textbf{Steps and rubric as graph}. We encode action steps and the corresponding scoring rubric with a directed acyclic graph (DAG). This DAG is denoted as $\mathcal{G}=(\mathbb{V}, \mathbb{E})$ with $\mathbb{V}$ as the set of nodes and $\mathbb{E}$ as the set of directed edges. $\mathbb{V}$ consists of three types of nodes: (1) the leaf nodes, denoted as ${V^s}$, correspond to individual action steps performed in the input video $X$; (2) the intermediate nodes capturing possible intermediate stages in the scoring criteria; and (3) a designated root node ${V^{r}}$ representing the final score of the action. Further, the edges $\mathbb{E}$ indicate the scoring rubric, connecting steps (leaf nodes) to stages (intermediate nodes), and stages (intermediate nodes) to the final score (root node). We note that $\mathcal{G}$ varies for every input video $X$ (assuming a single action), as different steps might be performed. \cref{fig:architecture} (a-b) show the example in diving where the key steps and scoring rubric are encoded using our DAG. Additional examples can be found in our supplement \textcolor{blue}{Fig.\ B}.

\smallskip
\noindent \textbf{Learning for quality assessment.} Our approach involves a two-step process for quality assessment. First, we employ an embedding function $f$, designed to map individual steps into a latent space representing action quality. Secondly, we leverage the key step embeddings $\{Z^s\}$ along with the score rubric encoded in $\mathcal{G}$ to learn a scoring function $h$. These functions $f$ and $h$ are defined as follows:
\begin{equation}
\small
f: X, \mathcal{G} \mapsto \{Z^s\}; \quad h: \{Z^s\}, \mathcal{G} \mapsto Y,
\end{equation}
where $\{Z^{s}\}$ are the embeddings for the set of steps $\mathbb{S}$ in $X$, corresponding to the leaf nodes $\{V^s\}$ on the DAG. 

\subsection{Modeling Score Uncertainty with Stochastic Embeddings}\label{sec:stochastic}

To model the prediction uncertainty, we adopt \textit{stochastic step embeddings} defined on the leaf nodes, such that $Z^s\in\mathbb{R}^D \sim p(Z^s | X, \{V^s\})$. Unlike deterministic embeddings, where $Z^s$ would be a fixed vector, stochastic embedding characterizes the distribution of $Z^s$, allowing for uncertainty control. Specifically, we model $p(Z^s | X, V^s)$ as a Gaussian distribution in $\mathbb{R}^D$ with mean $\mu^s$ and diagonal covariance $\Sigma^s$. The embedding function $f$ is thus tasked to predict the mean and covariance for the key steps $\mathbb{S}$, \ie, $\{\mu^s, \Sigma^s\} = f(X, \{V^s\})$.

\smallskip
\noindent \textbf{Propagating stochastic embeddings on the graph.} 
Our scoring function $h$ takes the stochastic embeddings ${Z^s}$ for leaf nodes in $\mathcal{G}$ (provided by $f$), further computes the embeddings for all nodes in $\mathcal{G}$, and finally decodes a quality score $Y$ from the embedding $Z^r$ of the root node $V^r$.
To this end, we propose an extension of graph neural networks (GNNs), in which stochastic embeddings ${Z^s}$ are propagated from leaf nodes ${V^s}$ to the root node $V^r$ based on the graph structured informed by the scoring rubric of a particular task. 
Key to this GNN lies in a lightweight MLP that operates on each node, taking as input the embeddings of its direct predecessors, and generating a new embedding that is further propagated to its successors.
This scoring function $h$ is thus given by
\begin{equation}
\small
\underbrace{
    Z^s \sim \mathcal{N}\left(\mu^s, \Sigma^s\right), \ \forall s \in \ \mathbb{S}
}_{\text{Sampling from leaf nodes}}; \quad 
\underbrace{
    Z^{\neg s} = G\left(\Sigma_{V^j \in \mathcal{P}(V^{\neg s})} Z^j\right)
}_{\text{Propagating on the DAG}}; \quad 
\underbrace{
    \hat{Y} = \mathcal{S}\left(Z^{r}\right)
}_{\text{Deocoding the score}}\label{eq:gnn}
\end{equation}
where $V^{\neg s}$ denotes a non-leaf node with its embedding $Z^{\neg s}$. $\mathcal{P}(V^{\neg s})$ is the set of predecessors of $V^{\neg s}$, $G(\cdot)$ is the MLP aggregating features from predecessors, and $\mathcal{S}({\cdot})$ is another MLP decoding the final score $\hat{Y}$ from the root node $V^r$. 

It is important to note that each leaf embedding $Z^s$ is stochastic, characterized by a Gaussian distribution ($p(Z^s | X, \mathcal{G}) = \mathcal{N}(\mu^s, \Sigma^s)$), with parameters predicted by $f$. The non-leaf embeddings are however deterministic given samples from leaf distributions. This design is motivated by our assumption of the scoring rubric, where uncertainty lies only in assessing action steps and identical individual scores will yield the same final score. %

\subsection{Learning with Variational Information Bottleneck} \label{sec:vib}

With stochastic embeddings, training of \name is a challenge. 
We design a training scheme under the variational information bottleneck framework. %

\smallskip
\noindent \textbf{Variational information bottleneck (VIB).} To train our model $p(Y|X, \mathcal{G})$ with stochastic step embeddings $\{Z^s\}$, we adopt the information bottleneck principle~\cite{tishby1999information}, leading to the maximization of the following objective
\begin{equation}
\small
    I(\{Z^s\}; Y | \mathcal{G}) \ -\ \beta I(\{Z^s\}; X | \mathcal{G}),
\end{equation}
where $I$ is the conditional mutual information, and $\beta>0$ controls the tradeoff between the sufficiency of using step embeddings $\{Z^s\}$ for predicting $Y$ given $\mathcal{G}$, and the size of the embeddings $\{Z^s\}$ derived from $X$ and $\mathcal{G}$. 

While mutual information is computationally intractable for high dimensional \{$Z^s$\}, a common solution~\cite{alemideep} is to assume Markov property ($p(Z|X, Y, \mathcal{G}) = p(Z|X, \mathcal{G})$) and conditional independence ($p(\{Z^s\} | X, \mathcal{G}) = \prod_s p(Z^s | X, \mathcal{G})$), followed by the variational approximation for a tractable lower bound
\begin{equation}
\small
    -\mathcal{L}_{\text{VIB}} = \mathbb{E}_{Z^s \sim p(Z^s | X,\mathcal{G}), \forall s \in \mathbb{S}} \left[ \log p(Y | \{Z^{s}\}, \mathcal{G}) \right] - \beta \Sigma_{s \in \mathbb{S}} \ \text{KL} \left( p(Z^s | X, \mathcal{G}) || p(Z^s | \mathcal{G}) \right),\label{eq:vib}
\end{equation}
where $p(Y | \{Z^s\}, \mathcal{G})$ is modeled by the scoring function $h$, $\text{KL}$ denotes the Kullback–Leibler divergence, and $p(Z^s | \mathcal{G})$ is an approximate marginal prior.

\smallskip
\noindent \textbf{VIB loss.}
The first term in~\cref{eq:vib} defines the log-likelihood of the score given the input. By assuming that output scores follow a Gaussian with a fixed variance $\sigma$, this term can be reduced to a mean squared error (MSE) loss
\begin{equation}
\small
    \mathcal{L}_{MSE} = \frac{1}{N}\Sigma_i^N (\hat{Y_i} - {Y_i})^2 / \sigma^2,
    \label{eq:mse}
\end{equation}
where $Y_i$ is the predicted score for a video indexed by $i$, $\hat{Y_i}$ is the corresponding ground-truth score, and $N$ is the total number of videos in the training set. 

The second term in~\cref{eq:vib} regularizes the latent space and encodes prediction uncertainty. By assuming a marginal prior of $\mathcal{N}(0, I)$ for $p(Z^s | \mathcal{G})$, we have
\begin{equation}
\small
    \begin{split}
    \mathcal{L}_{KL} &= \Sigma_{s \in \mathbb{S}} \  KL\left(\mathcal{N}(\mu^s (x),\Sigma^s(x) || \mathcal{N}(0, I) \right) \\
        &= \frac{1}{2} \Sigma_{s \in \mathbb{S}} \ \Sigma_{j=1}^D \left((\mu^s_j)^2 + (\sigma^s_j)^2 - \log(\sigma^s_j)^2 -1\right),
    \end{split}
\label{eq:kldiv} 
\end{equation}
where $\mu^s_j$ and $\sigma^s_j$, respectively, are the $j$-th dimension of the mean ($\mu^s(x)$) and variance (square root of the diagonal of $\Sigma^s(x)$), for the step $s$. 

The VIB loss $(\mathcal{L}_{VIB})$ is thus given by  
\begin{equation}
\small
    \mathcal{L}_{VIB} = \mathcal{L}_{MSE} + \beta \mathcal{L}_{KL}. 
    \label{eq:vib_loss}
\end{equation}
$\mathcal{L}_{VIB}$ consists of (a) the MSE loss $\mathcal{L}_{MSE}$ from the negative log-likelihood of the predicted scores, aiming at minimizing prediction errors; and (b) the KL divergence $\mathcal{L}_{KL}$ between the predicted Gaussian and the prior, regularizing the stochastic embeddings. Further, the coefficient $\beta$ balances between two loss terms.

During training, samples are drawn to compute the loss function. The output is matched to the Gaussian distribution with its mean equal to the average of the judge scores. The reparameterization trick~\cite{kingma2013auto} is used to allow the backpropagation of gradients through the sampling process.

\smallskip
\noindent \textbf{Estimating uncertainty.} The diagonal covariance $\Sigma^s(X)$ models the uncertainty of the predicted quality score of a step $s$ for an input video $X$. A larger value in its diagonal represents a wider distribution of scores and, hence, a lower confidence in the prediction. Following~\cite{oh2018modeling, li2021learning} we generate uncertainty scores by summing up the harmonic means of the predicted variances for individual steps
\begin{equation}
\small
    \text{uncertainty}(Y) = \Sigma_{s \in \mathbb{S}}{D/{\Sigma_{j=1}^D(\sigma_j^s)^{-1}}},
\end{equation}
where $D$ is the dimensionality of the stochastic embeddings. Again, $\sigma^s_j$ is the $j$-th dimension of the predicted variance.

\smallskip
\noindent \textbf{Stochastic vs.\ deterministic modeling}. An interesting variant of \name is to disable its stochastic component. Conceptually, this is equal to considering step embeddings $Z^r$ as vectors and removing the KL loss $\mathcal{L}_{KL}$. We refer to this deterministic version of our model as \namedet. Without stochastic embeddings, \namedet is unable to estimate prediction uncertainty, yet often yields slightly lower prediction errors. This trade-off is also observed in prior works~\cite{shi2019probabilistic,li2021learning,chun2021probabilistic}. We include this variant of our model in the experiments. 

\subsection{Model Instantiation and Implementation}

\smallskip
\noindent \textbf{Video and step representation.}
For an input video $X$, we adapt a pre-trained video backbone(\eg., I3D~\cite{carreira2017quo}) to extract its clip-level features, which are further pooled to produce video features $(x_1, x_2, ..., x_T)$ with fixed-length $T$. To represent action steps, we make use of a pre-trained language model~\cite{JMLR:v25:23-0870} %
(Flan-T5) to extract text features from their step descriptions, resulting in an ordered set of text embeddings $(s_1, s_2, ..., s_K)$ for $K$ steps. Note that the language model is not part of \name. It is used solely to extract embeddings for text descriptions of the action steps (see supplement \textcolor{blue}{Tables
I-L}).

\smallskip
\noindent \textbf{Embedding function $f$.} 
Our embedding function $f$ is realized using a Transformer model~\cite{vaswani2017attention} (see \cref{fig:architecture}(c)). $f$ first processes video features $(x_1, x_2, ..., x_T)$ with a self-attention block and text embeddings $(s_1, s_2, ..., s_K)$ using a MLP. It further makes use of cross-attention blocks (2x) to fuse video and text features, where video features are used to compute keys and values, and text embeddings of steps are projected into queries. Further, $f$ decodes stochastic embeddings of individual steps by predicting a mean vector $\mu^s \in \mathbb{R}^D$ and a diagonal covariance vector $\Sigma^s \in \mathbb{R}^D$ for each step $s$. 

\smallskip
\noindent \textbf{Scoring function $h$.} 
With Gaussian distributions for all steps specified by $\{\mu^s, \Sigma^s\}$, we encode the steps and score rubric into a video-specific DAG $\mathcal{G}$, and realize $h$ as a GNN defined on $\mathcal{G}$ following~\cref{eq:gnn}. $h$ is parameterized by its aggregation function $G$, which is shared among nodes of the same type. $G$ is implemented using an averaging operation followed by a MLP (2 layers). Finally, $h$ decodes the final score at the root note $V^r$ of $\mathcal{G}$. 

\smallskip
\noindent \textbf{Training with auxiliary losses.} While the VIB loss (\cref{eq:vib_loss}) is sufficient for training, it falls short of considering the temporal ordering of steps. This is because of the conditional independence assumption needed for the derivation of VIB, \ie, $p(\{Z^s\} | X, \mathcal{G}) = \prod_s p(Z^s | X, \mathcal{G})$, where the ordering of $\{Z^s\}$ is discarded. To bridge the gap, we incorporate an auxiliary loss term $\mathcal{L}_{Aux}$ inspired by~\cite{bai2022action}. Specifically, we re-purpose the last cross-attention map ($\mathbb{R}^{K \times T}$) from $f$ as a \textit{step detector}. This is done by computing a temporally-weighted center across the attention of each action step to every video time step (\ie, column-wise). We then enforce that (a) this center is co-located with the peak of the attention along video time steps using a sparsity loss~\cite{bai2022action}; and (b) all centers follow the temporal ordering of corresponding action steps using a ranking loss~\cite{bai2022action}. These two terms are summed up as the auxiliary loss, and further added to the VIB loss with a small weight $(0.1)$. In our ablation, we empirically verify that adding the auxiliary loss leads to a minor performance boost.

\smallskip
\noindent \textbf{Inference with sampling.} At the inference time, we enhance robustness by sampling 20 times and averaging their predictions to compute the final score.

\section{Experiments and Results}\label{sec:exp}

\smallskip
\noindent \textbf{Datasets.} Our evaluations are primarily reported on three publicly available benchmark datasets, namely FineDiving~\cite{xu2022finediving}, MTL-AQA~\cite{parmar2019and}, and JIGSAWS~\cite{gao2014jhu} in the main paper. In supplement \textcolor{blue}{Sec.\ B}, we also include results on the Cataract-101~\cite{cataract101} with cataract surgery videos.

\smallskip
\noindent \textbf{Evaluation metrics.} For all our experiments, we consider metrics on both the \emph{accuracy} of the prediction and the \emph{calibration} of the uncertainty estimates. 

\noindent \textbullet \ For \emph{accuracy}, we use two widely adopted metrics for AQA~\cite{tang2020uncertainty,bai2022action,xu2022finediving}, namely Spearman's rank correlation ($SRCC$) and relative L2 distance ($R\ell_{2}$). $SRCC$ measures how well the predicted scores are ranked w.r.t.\ the ground truth, while $R\ell_{2}$ summarizes the prediction errors. A model with more accurate predictions will have higher $SRCC$ and lower $R\ell_{2}$.

\noindent \textbullet \ For \emph{calibration}, we report the uncertainty versus error curve following~\cite{oh2018modeling,li2021learning}. To plot this curve, test samples are sorted by increasing uncertainty and divided into 10 equal-sized bins. The mean absolute error (MAE) is then computed for items in each bin. We also follow ~\cite{oh2018modeling,li2021learning} in employing Kendall's tau ($\tau$)~\cite{kendalltau}, a numerical measure ranging from -1 to 1 to quantify the correlation between the uncertainties and the prediction errors. A higher $\tau$ indicates better calibration, signifying that a model's uncertainty better aligns with prediction errors.

\smallskip
\noindent \textbf{Baselines.} \name is benchmarked against a set of strong baselines, including exemplar-free methods such as DAE~\cite{zhang2023auto}, USDL and MUSDL~\cite{tang2020uncertainty}, and exemplar-based ones such as CoRE~\cite{xu_core_2021}, TPT~\cite{bai2022action} and TSA~\cite{xu2022finediving}. We further include the deterministic version of our model \namedet, which trades the ability of uncertainty estimation for a minor boost in accuracy. Several baselines adopt a direct regression approach, without providing a confidence or uncertainty measure for predictions. USDL~\cite{tang2020uncertainty} and TPT~\cite{bai2022action} implicitly offer a confidence value. In these works, the probability of the predicted score bin serves as a proxy for uncertainty, computed as ($1.0-$confidence). DAE~\cite{zhang2023auto} outputs the standard deviation of the score distribution, which represents uncertainty.

We seek to ensure a fair comparison yet recognize that methods in our benchmark may consider different settings and/or various types of input. Most prior exemplar-free methods only consider a video as input. While \name does not utilize exemplars, it takes additional input of step information, \ie, step presence and their temporal ordering. On the other hand, previous exemplar-based methods also require the step information as used by \name, in addition to an input video and an exemplar database. Notably, step information is used to select exemplars, leading to improved results. For example, for diving videos, CoRE~\cite{xu_core_2021}, TPT~\cite{bai2022action} and TSA~\cite{xu2022finediving} use the diving number (DN) encoding steps and their ordering. Further, TSA~\cite{xu2022finediving} also requires the timing of individual steps during training. While it is infeasible to standardize the settings of all methods, we compare to the best reported results in our experiments.

\subsection{Results on FineDiving}
\noindent \textbf{Dataset.} FineDiving~\cite{xu2022finediving} is the largest public dataset for AQA, with 3000 video samples capturing various diving actions. The dataset covers 52 different action types, 29 sub-action types, and 23 difficulty degree types, providing a rich and diverse set of examples for AQA.
While this dataset contains temporal annotations for the steps, which can be used to improve the performance of AQA as demonstrated in~\cite{xu2022finediving}, we do not use these annotations for \name.

\smallskip
\noindent \textbf{Experiment setup.} %
We adhere to the experimental setup of the most recent baseline~\cite{xu2022finediving} using their train-test split, with 2251 videos for training and 749 videos for testing. 
We follow the input video settings used in~\cite{xu2022finediving} for \name and the baselines. Specifically, for each video, we uniformly sample $96$ frames, which are segmented into $9$ overlapping clips, each containing $16$ consecutive frames. We refer to supplement \textcolor{blue}{Sec.\ A.1} for further implementation details.

\begin{figure}[t!]
    \centering
        \begin{minipage}{.47\textwidth}
            \centering
    \includegraphics[width=0.8\linewidth]{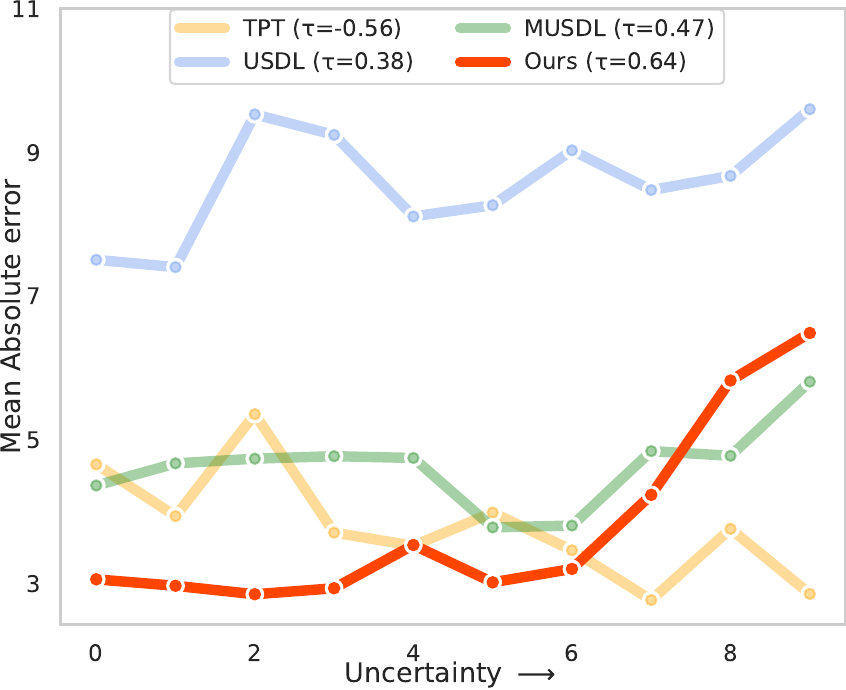}
            \subcaption{Uncertainty calibration on FineDiving}
            \label{fig:finediving_tau_results}
        \end{minipage}%
        \begin{minipage}{.47\textwidth}
            \centering           \includegraphics[width=0.8\linewidth]{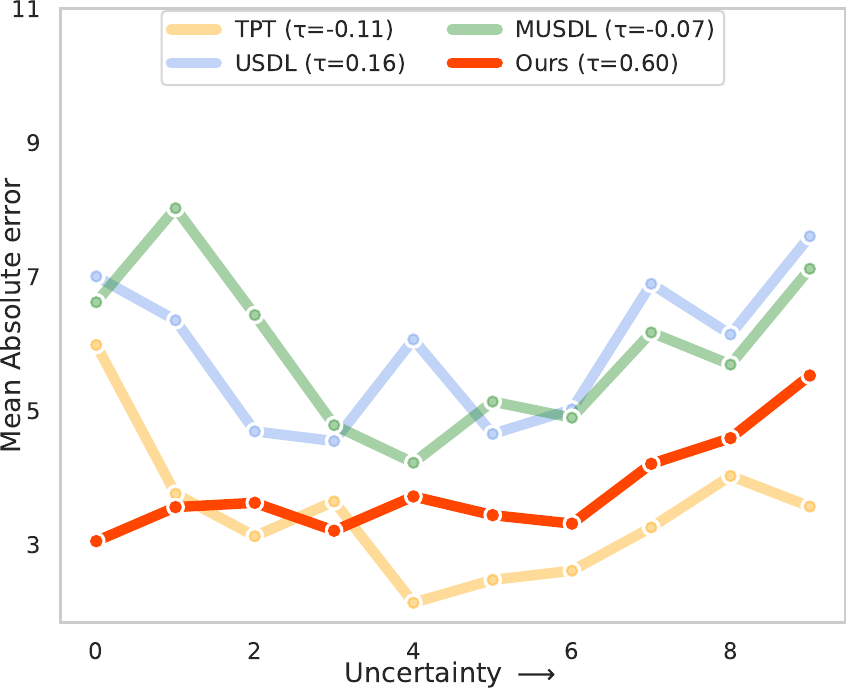}
            \subcaption{Uncertainty calibration on MTL-AQA}
            \label{fig:mtl_tau_results}
        \end{minipage}%
        \caption{\textbf{Uncertainty vs.\ prediction error (MAE)} on (a) Finediving and (b) MTL-AQA. Results are reported on the test splits, with the X-axis as the uncertainty bin index (uncertainty increases from left to right) and the Y-axis as the MAE in the bin. In comparison to baselines, \name has improved calibration with lower prediction errors.}%
\end{figure}

\smallskip
\noindent \textbf{Results.} %
\cref{tab:finediving_results} presents our results on FineDiving. Both our stochastic and deterministic versions (\name and \namedet) outperform the state-of-the-art TPT~\cite{bai2022action}, an exemplar-based method. \name shows a relative margin of 0.7\% / 1.4\% on $SRCC$ / $R\ell_{2}$, and \namedet has a relative margin of 0.9\% / 9.6\% on $SRCC$ / $R\ell_{2}$. This improvement is more pronounced when compared with the exemplar-free methods (MUSDL, DAE-MT) showcasing a significant relative gain of 4.9\%, 1.5\% on $SRCC$ and 29.8\%, 21.6\% on $R\ell_{2}$. While the deterministic \namedet has slightly higher accuracy, our stochastic \name demonstrates superior calibration of its uncertainty estimate ($\tau_{\text{\name}}=0.64$ vs.\ $\tau_{\text{TPT}}=-0.56$). 
\cref{fig:finediving_tau_results} further shows uncertainty calibration results. Uncertainty estimates from \name have a clear upward trend, indicating a higher calibration level. While MUSDL~\cite{tang2020uncertainty} also exhibits a reasonable level of calibration ($\tau_{\text{MUSDL}}=0.47$ vs.\ $\tau_{\text{\name}}=0.64$), the errors are significantly higher than \name across all uncertainty levels. 

\begin{figure*}[t]
    \centering
        \captionof{table}{\textbf{Main results} on (a) FineDiving and (b) MTL-AQA. Prediction accuracy ($SRCC$ and $R\ell_{2}$) and uncertainty calibration ($\tau$) metrics are reported. We compare our method with exemplar-based and exemplar-free baselines.}         
        \label{tab:combined_table_results}
    \begin{minipage}{\textwidth}
        \centering
        \begin{minipage}[t]{.5\textwidth}
        \subcaption{Results on FineDiving}
        \label{tab:finediving_results}
            \centering
            \resizebox{\columnwidth}{!}{
                \begin{tabular}{cccccc}
                    \hline %
                     & & & \multicolumn{3}{c}{\textbf{Metrics}}\\
                    \cline{4-6}
                    & & & \textbf{$SRCC(\uparrow)$} & \textbf{$R\ell_{2} (\downarrow)$}  & \textbf{$\tau (\uparrow)$}\\
                    \hline
                    \multirow{3}{*}{\shortstack{Exemplar\\based}} 
                    & \multicolumn{2}{c}{CoRe \cite{xu_core_2021}} & 0.9061 & 0.3615 & - \\
                    & \multicolumn{2}{c}{TSA \cite{xu2022finediving}} & 0.9203 & 0.3420 & -\\
                    & \multicolumn{2}{c}{TPT \cite{bai2022action}} & 0.9333     & 0.2877  & -0.5556 \\
                    \hdashline
                    \multirow{6}{*}{\shortstack{Exemplar\\free}}
                    & \multicolumn{2}{c}{USDL \cite{tang2020uncertainty,xu2022finediving}} & 0.8913 & 0.3822 & 0.3778 \\
                    & \multicolumn{2}{c}{MUSDL \cite{tang2020uncertainty,xu2022finediving}} & 0.8978 & 0.3704 & 0.4667\\
                    & \multicolumn{2}{c}{DAE\cite{zhang2023auto}} & 0.8820 &  0.4919 & -0.1999\\
                    & \multicolumn{2}{c}{DAE-MT \cite{zhang2023auto}} & 0.9285 & 0.3320 &  -0.4667 \\
                    \rowcolor{gray!15}
                    & \multicolumn{2}{c}{\name (Ours)} &  0.9402 & 0.2838  & \textbf{0.6444} \\ 
                    \rowcolor{gray!15}
                    & \multicolumn{2}{c}{\namedet (Ours)} &  \textbf{0.9421} &  \textbf{0.2600} & - \\ 
                    \hline
                \end{tabular}
             }
        \end{minipage}%
        \hfill
        \begin{minipage}[t]{.47\textwidth}
        \subcaption{Results on MTL-AQA}
        \label{tab:mtl_results}
            \centering
            \resizebox{\columnwidth}{!}{
        \begin{tabular}{cccccc}
        \hline
         &   &   & \multicolumn{3}{c}{\textbf{Metrics}}\\
        \cline{4-6}
         & & & \textbf{$SRCC (\uparrow)$} & \textbf{$R\ell_{2} (\downarrow)$} & \textbf{$\tau (\uparrow)$}\\
        \hline
        \multirow{4}{*}{\shortstack{Exemplar\\based}} 
        & \multicolumn{2}{c}{TSA-Net \cite{wang2021tsa}} &  0.9422 & - & - \\
        & \multicolumn{2}{c}{CoRe \cite{xu_core_2021}} & 0.9512 & 0.2600 & -\\
         & \multicolumn{2}{c}{DAE-CoRe \cite{zhang2023auto}} &  0.9589 & - & -\\
        & \multicolumn{2}{c}{TPT \cite{bai2022action}} &  0.9607 & 0.2378 & -0.1111\\
         
        \hdashline
        \multirow{7}{*}{\shortstack{Exemplar\\free}} 
        & \multicolumn{2}{c}{C3D-AVG-MTL \cite{parmar2019and}} & 0.9044 & - & -\\
         & \multicolumn{2}{c}{USDL \cite{tang2020uncertainty}} & 0.9231 & 0.4680 & 0.1556\\
         & \multicolumn{2}{c}{MUSDL \cite{tang2020uncertainty}} & 0.9273 & 0.4510 & -0.0667\\\
          & \multicolumn{2}{c}{DAE 
 \cite{zhang2023auto}} & 0.9231 &  - & -\\
         & \multicolumn{2}{c}{DAE-MT \cite{zhang2023auto}} & 0.9490 &  0.2738 & -0.4222\\
         \rowcolor{gray!15}
         & \multicolumn{2}{c}{\name (Ours)} & 0.9594 & 0.2580 & \textbf{0.6000} \\
         \rowcolor{gray!15}
          & \multicolumn{2}{c}{\namedet (Ours)} & \textbf{0.9620} & \textbf{0.2280}  & -\\
      \hline
    \end{tabular}
            }
        \end{minipage}
    \end{minipage}%
\end{figure*}

\subsection{Results on MTL-AQA}
\noindent \textbf{Dataset.} MTL-AQA~\cite{parmar2019and} is one of the most commonly used datasets for AQA. It consists of 1412 samples collected from 16 events with diverse views. The dataset has a rich set of annotations, including the steps performed during the dive, the difficulty score associated with the dive, and the individual judge scores.

\smallskip
\noindent \textbf{Experiment setup.} We follow the evaluation protocol of~\cite{xu_core_2021, bai2022action, xu2019learning}, dividing the dataset into the standard train set of 1059 videos and a test set of 353 videos. Further, we use the same input video settings as TPT~\cite{bai2022action} in our experiments to ensure a fair comparison. Specifically, for each video, we uniformly sample $103$ frames segmented into $20$ overlapping clips, each containing $8$ continuous frames. Please refer to supplement \textcolor{blue}{Sec.\ A.2} for more details.

\smallskip
\noindent \textbf{Results.} \cref{tab:mtl_results} summarizes our results on MTL-AQA. Similar to FineDiving, \name shows state-of-the-art results on MTL-AQA across all evaluation metrics. Specifically, our \namedet outperforms the best exemplar-free model (DAE-MT) by a relative margin of $1.4$\% / $16.7\%$ on $SRCC$ / $R\ell_{2}$. When compared with the competitive exemplar-based TPT, our has slightly better $SRCC$ ($SRCC_{TPT}=0.9607$ vs $SRCC_{\text{\namedet}}=0.9620$) and $R\ell_{2}$ (+$4.1\%$ relative margin). Again, compared to previous methods, our stochastic model shows improved calibration ($\tau_{\text{\name}}=0.60$ vs.\ $\tau_{USDL}=0.16$ vs.\ $\tau_{TPT}=-0.11$) as shown in~\cref{fig:mtl_tau_results}.

\subsection{Results on JIGSAWS}
\noindent \textbf{Dataset.} In addition to diving videos, we also evaluate \name on JIGSAWS~\cite{gao2014jhu} --- a robotic surgical video dataset.
The dataset includes three tasks: ``Suturing (S),'' with 39 recordings, ``Needle Passing (NP),'' with 26 recordings and ``Knot Tying (KT)'' with 36 recordings. JIGSAWS is widely used for action quality assessment, despite its small scale.
\begin{table}[t]
    \centering
    \footnotesize
    \caption{\textbf{Results on JIGSAWS~\cite{gao2014jhu} dataset}. Only prediction accuracy ($SRCC$) is considered due to the limited sample size. \name outperforms all prior approaches.}%
    \label{tab:jigsaw_results}
    \resizebox{0.5\columnwidth}{!}{
    \begin{tabular}{ccccccc}
        \hline
        &   &   & \multicolumn{4}{c}{\textbf{Task}}\\
        \cline{4-7}
         & & & \textbf{S} & \textbf{NP} & \textbf{KT} & \textbf{Avg} \\
        \hline
        \multirow{2}{*}{\shortstack{Exemplar\\based}} 
        & \multicolumn{2}{c}{CoRe \cite{xu_core_2021}} & 
        0.84 & 0.86 & 0.86 & 0.85\\
         & \multicolumn{2}{c}{TPT \cite{bai2022action}} &  
         0.88 & 0.88 & \textbf{0.91} & 0.89\\
        \hdashline
        \multirow{8}{*}{\shortstack{Exemplar\\free}} 
         & \multicolumn{2}{c}{ST-GCN \cite{yan2018spatial}} &  
         0.31 & 0.39 & 0.58 & 0.43\\
        & \multicolumn{2}{c}{TSN \cite{wang2016temporal}} &  
        0.34 & 0.23 & 0.72 & 0.46\\
        & \multicolumn{2}{c}{JRG \cite{pan2019action}} &  
        0.36 & 0.54 & 0.75 & 0.57\\
        
        & \multicolumn{2}{c}{USDL \cite{tang2020uncertainty}} & 
        0.64 & 0.63 & 0.61 & 0.63\\
         & \multicolumn{2}{c}{MUSDL \cite{tang2020uncertainty}} & 
         0.71 & 0.69 & 0.71 & 0.70\\
         & \multicolumn{2}{c}{DAE \cite{zhang2023auto}} & 
         0.73 & 0.72 & 0.72 & 0.72\\
         & \multicolumn{2}{c}{DAE-MT \cite{zhang2023auto}} & 
         0.78 & 0.74 & 0.74 & 0.76\\
          \rowcolor{gray!15}
           & \multicolumn{2}{c} {\namedet (Ours)} &  0.88 & 0.93	& 0.88 & 0.90\\
            \rowcolor{gray!15}
           & \multicolumn{2}{c} {\name (Ours)} &  \textbf{0.92}	& \textbf{0.94}	& \textbf{0.90} & \textbf{0.92}\\        
      \hline
    \end{tabular}
    }%
\end{table}

\smallskip
\noindent \textbf{Experiment setup.} Due to the limited number of samples in the dataset (as few as 7 videos in the test set), cross-validation is often considered for evaluation on JIGSAWS. To ensure a fair comparison, we follow the commonly adopted splits from~\cite{tang2020uncertainty}, and the input video setting from~\cite{bai2022action}. Specifically, for each video, we uniformly sample $160$ frames which are segmented into $20$ non-overlapping clips. We opt to not include score calibration curves due to the limited sample size of the test sets. Additionally, the key steps in JIGSAWS are general motions (\eg reaching for the needle, orienting the needle, etc.) and thus cannot be localized to any specific section of the video. Thus, we do not use the auxiliary losses for this experiment. More details are described in supplement \textcolor{blue}{Sec.\ A.3}. %

\smallskip
\noindent \textbf{Results.} 
\cref{tab:jigsaw_results} summarizes our results on JIGSAWS. Similar to previous datasets, our models exhibit notable advancements over the previous state-of-the-art model TPT~\cite{tang2020uncertainty}, showcasing substantial improvements of 1.1\% (\name) to 3.4\% (\namedet) in terms of average $SRCC$ relative to the exemplar-based state-of-the-art TPT~\cite{bai2022action}. When compared to the exemplar-free methods, our approach demonstrates an impressive 18.4\%  (\name) to 21.0\% (\namedet) relative gain in average $SRCC$ compared to the latest method DAE-MT~\cite{zhang2023auto}.

\subsection{Ablation Studies}
\label{sec:ablations}
To understand our model design choices, we conduct ablation studies on the MTL-AQA~\cite{parmar2019and} dataset.
Additional ablations are in supplement \textcolor{blue}{Sec.\ C.2}.

\smallskip
\noindent \textbf{Experiment setup.} To simplify our experiments, we opt for running our ablations using fixed I3D features. This allows us to precisely evaluate the contribution of different components of our model. Specifically, we choose I3D weights from an intermediate checkpoint of our trained model and extract features for all videos with the frozen backbone. 

\smallskip
\noindent  \textbf{Base model.} Our ablation constructs a base model using \textit{randomly initialized} step embeddings, an averaging of these embeddings after cross attention with the video features, followed by an MLP for scoring. This base model is trained using only the MSE loss. We then gradually add modules from \name and study their effects. \cref{tab:mtl_ablations} presents our results using the same features and training epochs, with our base model in row 1.

\begin{table}[t]
\centering
\small
\caption{\textbf{Ablation studies} of model components on MTL-AQA dataset. 
* indicates that the text embeddings were frozen during training.}%
\label{tab:mtl_ablations}
\resizebox{0.85\columnwidth}{!}
{%
\setlength{\tabcolsep}{4pt}
\begin{tabular}{cccccccc}
\hline
\multirow{2}{*}{\textbf{\begin{tabular}[c]{@{}c@{}}Step\\ Rep.\end{tabular}}} & \multirow{2}{*}{\textbf{\begin{tabular}[c]{@{}c@{}}DAG\\ (Rubric)\end{tabular}}} & \multirow{2}{*}{$\boldsymbol{\mathcal{L}_{KL}}$} & \multirow{2}{*}{$\boldsymbol{\mathcal{L}_{Aux}}$} & \multicolumn{4}{c}{\textbf{Metrics}} \\ \cline{5-8} 
 &  &  &  & $\boldsymbol{SRCC}(\uparrow)$ & $\boldsymbol{R\ell_2} (\downarrow)$ & $\boldsymbol{\tau} (\uparrow)$ & \textbf{Avg.\ Rank} ($\downarrow$) \\ \hline
Random & $\times$ & $\times$ & $\times$ & 0.9426 & 0.3882 & - & 5.50 \\
Text & $\times$ & $\times$ & $\times$ & 0.9431 & 0.3509 & - & 4.50 \\
Text & $\checkmark$ & $\times$ & $\times$ & 0.9430 & 0.3336 & - & 4.50 \\
Text* & $\checkmark$ & $\times$ & $\times$ & 0.9437 & 0.3335 & - & 3.50 \\
Text* & $\checkmark$ & $\checkmark$ & $\times$ & 0.9448 & 0.3329 & 0.4222 & 1.83 \\
Text* & $\checkmark$ & $\checkmark$ & $\checkmark$ & \textbf{0.9460} & \textbf{0.3303} & \textbf{0.4222} & \textbf{1.17} \\ \hline
\end{tabular}
}%
\end{table}

\smallskip 
\noindent \textbf{Text embeddings as step representations.} We first replace randomly initialized step representations with the text embeddings of step descriptions. This leads to a major boost in $R\ell_{2}$ (\cref{tab:mtl_ablations} row 1 vs.\ row 2), by leveraging knowledge encoded in the LLM~\cite{JMLR:v25:23-0870}. Further, we find that freezing the text embeddings leads to comparable results and faster convergence (\cref{tab:mtl_ablations} row 3 vs.\ row 4).

\smallskip
\noindent \textbf{Does the scoring rubric help?} We also investigate the effects of encoding steps and rubric as a DAG---a key design of our model. Adding the DAG results in a noteworthy boost in $R\ell_{2}$ (\cref{tab:mtl_ablations} row 2 vs.\ row 3). This improvement can be ascribed to the DAG's proficiency in dissecting the action quality across steps.

\smallskip
\noindent \textbf{Effects of loss functions.}
We now study the loss terms. Our loss function has three terms (a) the MSE loss ($\mathcal{L}_{MSE}$) to minimize prediction error, (b) the KL loss ($\mathcal{L}_{KL}$) to regularize the stochastic embeddings, and (c) the auxiliary loss ($\mathcal{L}_{Aux}$) to ensure temporal ordering of steps. Adding the KL loss $\mathcal{L}_{KL}$ yields similar results in $SRCC$ and $R\ell_{2}$, yet enables calibrated uncertainty estimation. Further attaching the auxiliary loss $\mathcal{L}_{Aux}$ leads to improvement in both $SRCC$ and $R\ell_{2}$, while maintaining the calibration performance.

\begin{figure*}[t]
    \centering
    \begin{subfigure}{\textwidth}
    \centering
        \includegraphics[width=0.99\linewidth]{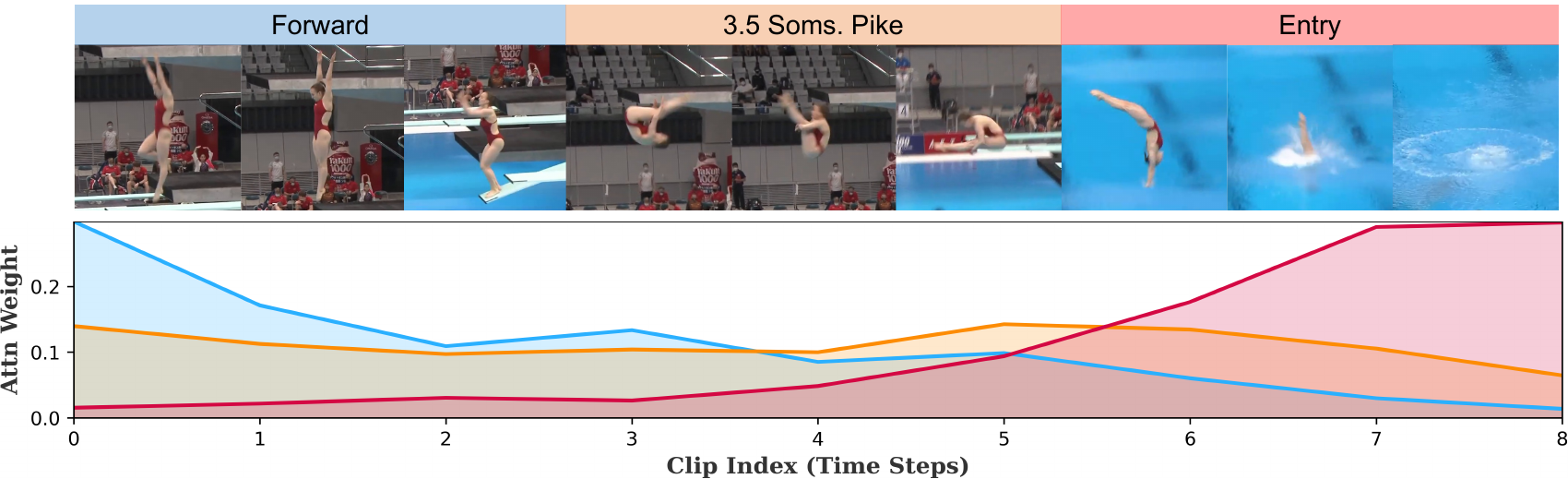}
    \end{subfigure}
    \begin{subfigure}{\textwidth}
    \centering
        \includegraphics[width=0.99\linewidth]{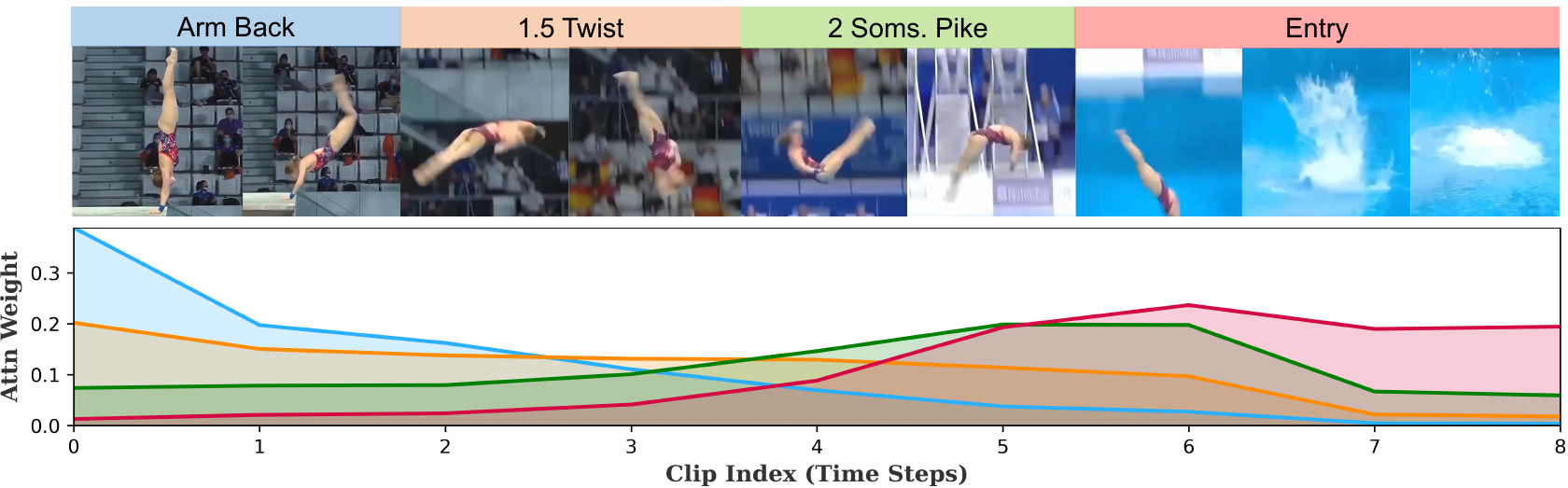}
    \end{subfigure}
    \caption{\textbf{Visualization of the cross-attention maps}. Y-axis : attention value; Y-axis: clip indices (time).
    Each curve shows an attention map from a step representation to the temporal video features. The frames shown above are aligned with the timing of the corresponding attention plot. Curves and steps are colored accordingly. %
    } \label{fig:cross_attention_plots_visual}
\end{figure*}

\smallskip
\noindent \textbf{Evaluating the cross-attention maps.}
To gain insight into \name, we now examine the cross-attention maps between the step representations and video features in our learned embedding function $f$. \cref{fig:cross_attention_plots_visual} visualizes the attention map on two test videos on FineDiving. These maps reveal that a step representation is likely to attend to video features during which the step occurs, indicating that \name learns to encode the temporal location of individual steps. 

We further evaluate this \emph{localization} ability following the Pointing game protocol~\cite{zhangTopDownNeuralAttention2018}, widely considered in weakly supervised / unsupervised localization tasks~\cite{pointingame_cite1,pointingame_cite2}. Pointing Game compares a generated heatmap with an annotated time interval and counts the chance of the heatmap's peak falling into the specified interval. Our evaluation focuses on the FineDiving dataset since it is the only dataset providing annotated time intervals for individual steps. When evaluated on the full test set, attention maps from our model attain an accuracy of $61.4\%$ in the Pointing game protocol, significantly outperforming the chance level accuracy of $30.7\%$ (given each video has $3.26$ steps on average). Note that we did not use any annotated segmentation data for training.

\section{Conclusion and Discussion}
In this paper, we present a deep probabilistic model for action quality assessment in videos. Our key innovation is to integrate score rubrics and to model prediction uncertainty. Specifically, we propose to adapt stochastic embeddings to quantify the uncertainty of individual steps, and to decode action scores using a variant of graph neural network operating on a DAG encoding the score rubric. Our method offers an exemplar-free approach for AQA, achieves new state-of-the-art results in terms of prediction accuracy on public benchmarks, and demonstrates superior calibration of the output uncertainty estimates. We believe that our work provides a solid step towards AQA. 
We hope that our method and findings can shed light on the challenging problem of trustworthy video recognition.

\smallskip
{\noindent{\bf Acknowledgement}:
This work was supported by the UW Madison Office of the Vice Chancellor for Research with funding from the Wisconsin Alumni Research Foundation, by National Science Foundation under Grant No.\ CNS 2333491, and by the Army Research Lab under contract number W911NF-2020221.}

\section*{Supplement}  %

\setcounter{section}{0}  %
\renewcommand{\thesection}{\Alph{section}}  %
\renewcommand{\thesubsection}{\thesection.\arabic{subsection}}  %

\setcounter{figure}{0}  %
\setcounter{table}{0}  %
\renewcommand{\thefigure}{\Alph{figure}}  %
\renewcommand{\thetable}{\Alph{table}}  %

In this supplement, we describe (1) technical, implementation, and experiment details for individual datasets, as well as our loss function (Sec.\ \ref{sec:impl_details}); (2) additional results for surgical skill assessment on Cataract-101 dataset (Sec.\ \ref{sec:cataract}); (3) additional ablations including the study of architectural design and loss coefficients, choice of text embeddings and additional cross-attention plots, and the consideration of action recognition methods (Sec.\ \ref{sec:ablations}); (4) details of the scoring rubric considered in the model, with examples from FINA diving manual (Sec.\ \ref{sec:rubric}); and (5) further discussion of our work (Sec.\ \ref{sec:discussion}). We hope this document complements our paper. 

For sections, figures, and tables, we use numbers (\eg, Sec.\ 1) to refer to the main paper and capital letters (\eg, Sec.\ A) to refer to this supplement.

\section{Technical, Implementation, and Experiment Details}\label{sec:impl_details}
We describe implementation and experiment details for 4 datasets considered in our paper, including FineDiving~\cite{xu2022finediving}, MTL-AQA~\cite{parmar2019and}, JIGSAWS~\cite{gao2014jhu}, and Cataract-101~\cite{cataract101}. We further present technical details of our loss functions. 

\subsection{Details for FineDiving} \label{sec:supp_finediving}
We follow the implementation from TSA~\cite{xu2022finediving} to process the input videos. Specifically, we uniformly sample $96$ frames from each video, segmented into $9$ overlapping clips of $16$ consecutive frames, with a stride of $10$. The frames are resized to a resolution of $200\times112$. During training, we employ a random crop of $112\times112$, while a center crop of size $112\times112$ is performed during testing. 

We use the I3D backbone~\cite{carreira2017quo} to extract video features, following prior works~\cite{bai2022action,xu2022finediving}. We generate text descriptions of steps with the help of GPT-4~\cite{openai2023gpt4}, as shown in Table~\ref{tab:fine_aqa_textqueries}. These descriptions are embedded using Flan-T5 XXL~\cite{flan_xxl_hugging_face, flan_xxl_paper}.

For training our model, we utilize AdamW~\cite{adamw}
optimizer with a linear warmup for $5$ epochs and a total of $350$ epochs. The training batch size is set to $8$, while the learning rates for the I3D backbone, transformer blocks, and the head (DAG) are set to $1\times10^{-5}$,  $3\times10^{-5}$, and  $5\times10^{-4}$, respectively. We also experimented with learning rate decay, yet did not find it helpful with our long training schedule. 

\subsection{Details for MTL-AQA} \label{sec:supp_mtl_aqa}
We follow TPT~\cite{bai2022action} for processing the input videos. Specifically, we uniformly sample $103$ frames from each video, segmented into 20 overlapping clips of 8 consecutive frames with a stride of 5. The frames are resized to a resolution of $455\times256$. During training, we employ a random crop of $224\times224$, while a center crop of size $224\times224$ is performed during testing.

Again, we use the I3D backbone~\cite{carreira2017quo} to extract video features, generate text descriptions of individual steps with the help of GPT-4~\cite{openai2023gpt4} (see Table~\ref{tab:mtl_aqa_textqueries}), and further embed these descriptions using Flan-T5 XXL~\cite{flan_xxl_hugging_face, flan_xxl_paper}

For training, we use the AdamW~\cite{adamw} optimizer with a linear warmup for $5$ epochs and a total of $350$ epochs. Other hyperparameters are kept the same as for FineDiving (batch size of $8$ with learning rates for I3D backbone, transformer blocks, and the head (DAG) as $1\times10^{-5}$,  $3\times10^{-5}$, and  $5\times10^{-4}$, respectively).

\subsection{Details for JIGSAWS} \label{sec:supp_jigsaws}
We adopt the video input configuration from TPT~\cite{bai2022action}. We uniformly sample $160$ frames from each video, and these frames are split into $20$ non-overlapping clips of $8$ consecutive frames with a stride of $8$. The frames are resized to a resolution of $455\times288$. During training, we employ a random crop of size $224\times224$, while during testing, a center crop of the same size ($224\times224$) is performed.

We follow the same protocol to extract video features (using I3D) and step embeddings (using Flan-T5 XXL), as FineDiving and MTL-AQA experiments. Step descriptions generated with the help of GPT-4 are shown in Table~\ref{tab:jigsaws_aqa_textqueries}.

For training, we utilize AdamW~\cite{adamw} optimizer with a linear warmup for the initial $5$ epochs and a total of $350$ epochs. Due to the smaller size of the dataset, a small training batch size of $2$ is employed. The learning rates for the I3D backbone, transformer blocks, and the head (DAG) are set to $1\times10^{-5}$, $3\times10^{-5}$, and $5\times10^{-4}$, respectively, maintaining consistency with the previous settings.

We note that in the JIGSAWS dataset, the steps are not performed in any specific order. Moreover, steps are repeated multiple times within each video. Consequently, we do not employ the auxiliary losses ($\mathcal{L}_{Aux}$) of ranking and sparsity for our experiments on JIGSAWS.

\subsection{Cataract-101} \label{sec:supp_cataract}
Videos in this dataset record complex surgical procedures (cataract surgery) and are thus significantly longer than other datasets (10 minutes vs.\ a few seconds). Consequently, we consider a stronger video backbone --- SlowFast-$8\times8$-R50~\cite{feichtenhofer2019slowfast}. We split an input video into sliding windows (clips) of $64$ frames with a temporal stride of $16$ frames. For each clip, we resize all frames such that their shortest side is kept at $256$. After a center-crop of size $224\times224$, the clips are fed into the SlowFast model pre-trained on Kinetics. The extracted video features are further used as input to our model for training and inference. It is worth noting that we do not update the SlowFast model during training, and thus, these video features remain fixed.  

We follow the same protocol to extract step embeddings (using Flan-T5 XXL), with step descriptions in Table~\ref{tab:cataract_aqa_textqueries}. We utilize AdamW~\cite{adamw} optimizer with a linear warmup for $5$ epochs and a total of $350$ epochs for training. The batch size is $2$ and the learning rate is $1\times10^{-4}$. 

\subsection{Details of Loss Functions}
As discussed in Sec.\ 3.3 and 3.4, our loss function consists of the VIB loss ($\mathcal{L}_{\text{VIB}}$) and the auxiliary loss ($\mathcal{L}_{\text{Aux}}$). Further, $\mathcal{L}_{\text{VIB}}$ comprises the MSE loss ($\mathcal{L}_{MSE}$) and the KL loss ($\mathcal{L}_{KL}$), and $\mathcal{L}_{\text{Aux}}$ includes the ranking loss ($\mathcal{L}_{rank}$) and the sparsity loss ($\mathcal{L}_{sparsity}$). Our overall loss is thus given by 
\begin{equation}
\small
    \mathcal{L} = \mathcal{L}_{MSE} + \beta \mathcal{L}_{KL} + \gamma \left( \mathcal{L}_{rank} + \mathcal{L}_{sparsity} \right),
\end{equation}
where $\beta$ and $\gamma$ are the loss weights. $\beta$ controls the bottleneck effect, and $\gamma$ manages additional regularization (\eg, temporal ordering).

The auxiliary losses of ranking ($\mathcal{L}_{rank}$) and sparsity ($\mathcal{L}_{sparsity}$) are designed to enforce the ordering of the step representation. Specifically, we follow TPT~\cite{bai2022action} and consider the last cross-attention map from our embedding function $f$. Concretely, given $S$ step representations as the queries, \ie $Q \in \mathbb{R}^{S\times D}$, and video features defined over $T$ time steps as the keys, \ie $K \in \mathbb{R}^{T \times D}$, we denote their cross-attention map as $A \in \mathbb{R}^{S\times T}$. Considering the cross-attention map for each step (\ie a row in $A$), a corresponding temporally-weighted center ($\Bar{\alpha_s}$), as a detector of the step's temporal location, is calculated as
\begin{equation}
\small
    \Bar{\alpha}_s = \Sigma_{t=1}^T t\cdot A_{s,t},
    \label{eq:center}
\end{equation}
where $\alpha_{s,t}$ is the similarity between a step representation and a video feature at time step $t$. Our auxiliary losses are defined on top of these centers.

The sparsity loss $\mathcal{L}_{sparsity}$ is defined to discourage the spread of query attention densities. $\mathcal{L}_{sparsity}$ is given by
\begin{equation}
\small
         L_{sparsity} = \Sigma_{s=1}^{S} \Sigma_{t=1}^{T} |t - \Bar{\alpha}_{s}| \cdot \alpha_{s,t}
\end{equation}

The ranking loss $\mathcal{L}_{rank}$ is designed to encourage all centers to follow the pre-specified step ordering. $\mathcal{L}_{rank}$ is written as
\begin{equation}
\small
    L_{rank} =\Sigma_{s=1}^{S-1} \max(0, \Bar{\alpha}_s-\Bar{\alpha}_{s+1} + m) + \max (0, 1- \Bar{\alpha}_1 + m) + \max(0, \Bar{\alpha}_S - T + m) 
\end{equation}

\section{Results on Cataract-101}\label{sec:cataract}
We now present our results on Cataract-101, a video dataset for surgical skill assessment. These results are omitted from the main paper due to lack of space.\medskip

\noindent \textbf{Dataset.} Cataract-101~\cite{cataract101} is a publicly available dataset of 101 cataract surgery videos, with each recorded procedure annotated with frame-level labels and the performing surgeon's corresponding skill score. The skill scores are discrete, labeled as either expert or novice.
\medskip

\noindent \textbf{Experiment setup.} 
We train our model on 80 randomly picked videos and test on the remaining 21 videos. Due to the significantly longer duration of videos (exceeding 10 minutes compared to a few seconds for diving videos), we perform our experiments with pre-extracted features. In this case, we choose a more advanced model for feature extraction. Specifically, we utilize SlowFast~\cite{feichtenhofer2019slowfast} model pre-trained on Kinetics~\cite{kay2017kinetics} to extract features for the videos and train our model on the extracted features. Given the binary labels, we switch from the mean squared error (MSE) loss to binary cross entropy (BCE) loss, which is compatible with our maximum log-likelihood interpretation of the loss. 

We report the average accuracy as our evaluation metric and compare our results to a baseline method TUSA~\cite{unifiedskill} specifically designed for surgical skill assessment. TUSA is trained and evaluated using the same features as our method.\medskip

\noindent \textbf{Results and discussion.} Both TUSA and our method reach an impressive 100\% accuracy. Randomizing train-test splits leads to similar perfect accuracy. Our analysis of this dataset shows the duration of the surgery is a good predictor of the surgeon's skills; expert surgeons often perform this routine surgery faster than novice surgeons.

\section{Additional Ablation Studies}\label{sec:ablations}
We further discuss additional ablation studies. Similar to the ablations in the main paper (Sec.\ 4.4), we run the experiments on the MTL-AQA~\cite{parmar2019and} dataset with the same pre-extracted I3D features unless otherwise specified.

\begin{table}[t]
\caption{\textbf{Ablation on loss weight $\beta$.} We report the accuracy and calibration metrics.}
\centering
\footnotesize
\begin{tabular}{cccccc}
\hline
 & \multicolumn{3}{c}{\textbf{Metrics}} \\ \cline{2-6} 
\boldsymbol{$\beta$} & \textbf{$SRCC (\uparrow)$} & \textbf{$R\ell_2 (\downarrow)$} & \textbf{$\tau (\uparrow)$} \\ \hline
$10^{-1}$ & 0.9450 & 0.3614 & 0.5556 \\
$10^{-2}$ & 0.9456 & 0.3529 & 0.4667 \\
$10^{-3}$ & 0.9463 & 0.3478 & 0.3333 \\
$10^{-4}$ & 0.9451 & 0.3289 & 0.3333 \\ 
$10^{-5}$ & 0.9449 & 0.3393 & 0.4222 \\ 
\hline
\end{tabular}
\label{tab:beta_ablations}%
\end{table}

\subsection{Effects of Loss Coefficients}
We first evaluate the effects of loss weights $\beta$ and $\gamma$. To this end, we fix $\gamma$ and vary $\beta$, which balances the objectives of minimizing prediction error and ensuring uncertainty estimation.  Table~\ref{tab:beta_ablations} shows the results. Lower values of $\beta$ often lead to minorly improved accuracy metrics, yet higher values of $\beta$ allow better calibration. Through the experiments, we empirically observe that an optimal balance can be attained with a training schedule in which $\beta$ is initially set to $10^{-5}$ and incrementally increased to a maximum value of $\beta=0.005$ during the training process. All results for \name in our study are reported using this annealing scheme, which was also discussed in prior work~\cite{kingma2013auto}. 

We further experiment with different values for $\gamma$, which denotes the weight for the auxiliary losses of ranking and sparsity. We empirically find our results are insensitive to different values and set $\gamma=0.1$ for our experiments.

\begin{table*}[t]
\centering
\caption{\textbf{Ablation on design choices}. We vary the design of major components in \name and report the results on MTL-AQA. \#Convs denotes the number of 1D convolution blocks, \#Enc denotes number of encoder blocks and \#Dec denotes the number of decoder blocks}
\label{tab:arch_ablation_table}
\resizebox{\textwidth}{!}{%
\begin{tabular}{cccccccccccc}
\hline
\multirow{2}{*}{\textbf{\begin{tabular}[c]{@{}c@{}}Embed\\ Dim\end{tabular}}} & \multirow{2}{*}{\textbf{\#Convs}} & \multirow{2}{*}{\textbf{\#Enc}} & \multirow{2}{*}{\textbf{\#Dec}} & \multirow{2}{*}{\textbf{\begin{tabular}[c]{@{}c@{}}DAG\\ (Rubric)\end{tabular}}} & \multirow{2}{*}{$\boldsymbol{\mathcal{L}_{spar}}$} & \multirow{2}{*}{$\boldsymbol{\mathcal{L}_{rank}}$} & \multirow{2}{*}{$\boldsymbol{\mathcal{L}_{KL}}$} & \multicolumn{4}{c}{\textbf{Metrics}} \\ \cline{9-12} 
 &  &  &  &  &  &  &  & $\boldsymbol{SRCC}(\uparrow)$ & $\boldsymbol{R\ell_2} (\downarrow)$ & $\boldsymbol{\tau} (\uparrow)$ & \textbf{Avg. Rank} ($\downarrow$) \\ \hline
256 & 2 & 0 & 0 & $\times$ & $\times$ & $\times$ & $\times$ & 0.9267 & 0.4213 & - & 6.67 \\
512 & 2 & 2 & 0 & $\times$ & $\times$ & $\times$ & $\times$ & 0.9263 & 0.3843 & - & 6.33 \\
512 & 2 & 2 & 0 & $\times$ & $\times$ & $\times$ & $\times$ & 0.9272 & 0.4090 & - & 6.00 \\
512 & 2 & 2 & 2 & $\times$ & $\times$ & $\times$ & $\times$ & 0.9444 & 0.3707 & - & 4.33 \\
512 & 2 & 2 & 2 & $\checkmark$ & $\times$ & $\times$ & $\times$ & 0.9437 & 0.3335 & - & 4.00 \\
512 & 2 & 2 & 2 & $\checkmark$ & $\checkmark$ & $\times$ & $\times$ & 0.9440 & 0.3562 & - & 4.33 \\
512 & 2 & 2 & 2 & $\checkmark$ & $\checkmark$ & $\checkmark$ & $\times$ & 0.9451 & 0.3458 & - & 3.33 \\
512 & 2 & 2 & 2 & $\checkmark$ & $\checkmark$ & $\checkmark$ & $\checkmark$ & \textbf{0.9460} & \textbf{0.3303} & \textbf{0.4222} & \textbf{1.00} \\ \hline
\end{tabular}%
}
\end{table*}

\subsection{Design Choices}
We vary the design of our model and examine their contributions to final performance. Specifically, our model extracts video features and step embeddings, further encodes individual features, uses cross-attention Transformer blocks to fuse them and decode step representation, and finally decodes a distribution of scores using a DAG representing the rubric. We study the design of encoding (convolution and self-attention) and decoding modules (cross-attention), as well as the choice of loss terms. 

Table~\ref{tab:arch_ablation_table} shows the results.  In addition to what is presented in the main paper, we highlight some key choices to boost performance in Table~\ref{tab:arch_ablation_table}: (a) using \emph{embedding dim} of $512$ over $256$ and utilizing $2$ decoder blocks over $1$ decoder block; and (b) combining all loss terms. 
We can clearly notice our full model (shown in the last row of Table \ref{tab:arch_ablation_table}) leads to the best performance in accuracy and score calibration. Specifically, this model combines two 1D-conv blocks, $2$ self-attention blocks, integration of text queries, and $2$ decoder blocks, followed by the DAG, and is optimized with $\mathcal{L}_{KL}$ and the auxiliary losses.

\subsection{Effects of Text Queries}
To assess the influence of various language models, we leverage text embeddings derived from three different models: \emph{OpenClip}~\cite{ilharco2021openclip}, $E5_{large}$~\cite{e5_wang2022text}, and \emph{Flan-T5 XXL}~\cite{flan_xxl_hugging_face, flan_xxl_paper}. The corresponding results are detailed in Table \ref{tab:text_query_ablation}. Our empirical findings indicate that embeddings derived from \emph{Flan-T5 XXL}~\cite{flan_xxl_hugging_face, flan_xxl_paper} result in the best performance on accuracy metrics  ($SRCC$ and $R\ell_2$). 

\begin{table}[t]
\caption{\textbf{Ablation on text embeddings.} We experiment with different language models and report results on MTL-AQA. \emph{Dim} denotes the text embedding dimensions.}     \label{tab:text_query_ablation}
    \setlength{\tabcolsep}{6pt}
    \centering
    \footnotesize
\begin{tabular}{cccccc}
    \hline
    \multirow{2}{*}{\textbf{Text Model}}  & \multirow{2}{*}{\textbf{Dim}}  &\multicolumn{3}{c}{\textbf{Metrics}}\\
    \cline{3-6}
      &  & \textbf{$SRCC (\uparrow)$} & \textbf{$R\ell_2 (\downarrow)$} & \textbf{$\tau (\uparrow)$}\\
    \hline
     OpenClip~\cite{ilharco2021openclip} & 512 &  0.9455 & 0.4339 & 0.2444\\
     E5$_{large}$~\cite{e5_wang2022text} &  1024 & 0.9431 & 0.3410 &  \textbf{0.4222} \\
     Flan-T5 XXL~\cite{flan_xxl_paper} & 4096 &  \textbf{0.9460}  & \textbf{0.3303} & \textbf{0.4222}\\       
  \hline
\end{tabular}%
\end{table}

\begin{figure*}[t]
    \centering
    \includegraphics[width=0.95\linewidth]{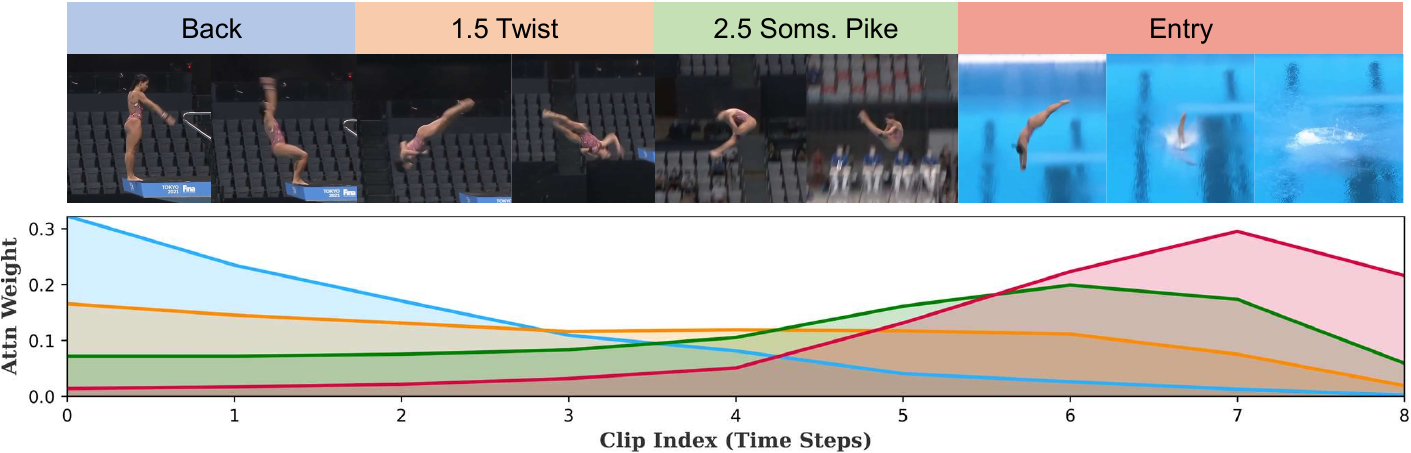}
    \includegraphics[width=0.95\linewidth]{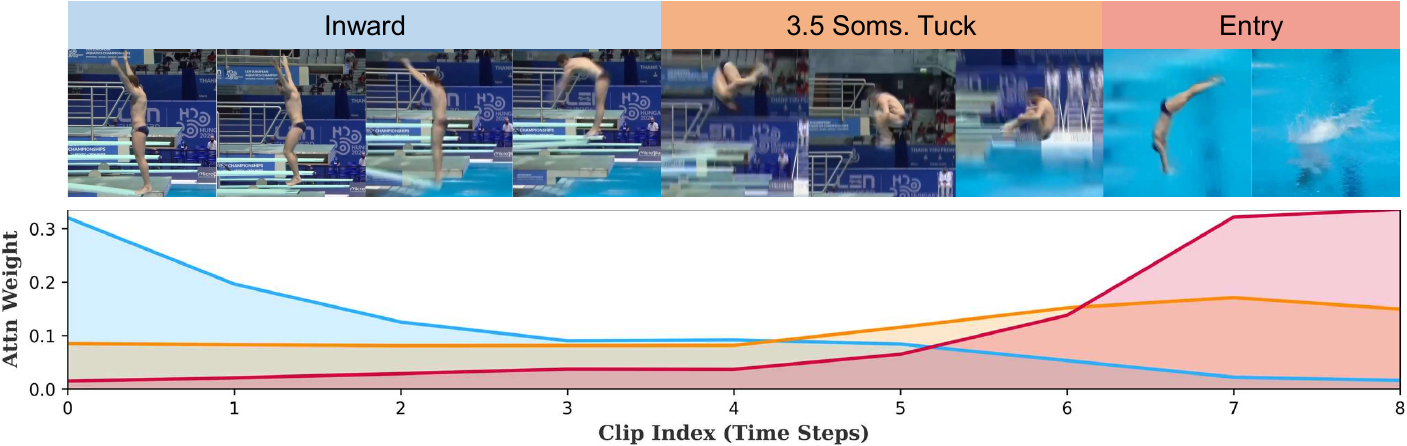}   \caption{\textbf{Visualization of the cross-attention maps} on two test videos from FineDiving~\cite{xu2022finediving} with the Y-axis as attention value and the X-axis as clip indices (time). Each curve shows an attention map from a step representation to the temporal video features. The frames shown are aligned with the time axis of the corresponding attention plot. Curves and steps are colored accordingly}
    \label{fig:cross_attention_plots}%
\end{figure*}

\subsection{What if Steps Are Not Available?}
When step information is not available at inference time, action recognition methods offer a solution for identifying steps in the input video. To explore this scenario, we experimented on the FineDiving dataset~\cite{xu2022finediving}, utilizing pre-extracted I3D features. For this task, we designed a simple action recognition model comprising a 2-layer MLP projection block, 2 Transformer blocks, and 2 MLPs. This model was trained on the training split to predict the dive number (\ie the steps in a video). Subsequently, we evaluated the trained model on the test set, achieving an average step recognition accuracy of $82\%$. The generated step predictions on the test set were then incorporated into \name, replacing the ground truth step presence information. Using the model-predicted step information resulted in an $SRCC$ and $R\ell_{2}$ of 0.9379 and 0.2779 respectively while the ground truth $SRCC$ and $R\ell_{2}$ are 0.9389 and 0.2750. These results demonstrate the feasibility of leveraging action recognition methods to provide step information for \name during inference.

\subsection{Additional Visualization of Our Results}
We present additional visualization of sample results in Fig.\ \ref{fig:cross_attention_plots}. These samples are from the test set of FineDiving, and the visualization follows the same format as Fig.\ 4 of our main paper.

\section{Score Rubric and Our DAG Representation}\label{sec:rubric}

\begin{figure*}[t]
    \centering
    \includegraphics[width=0.95\linewidth]{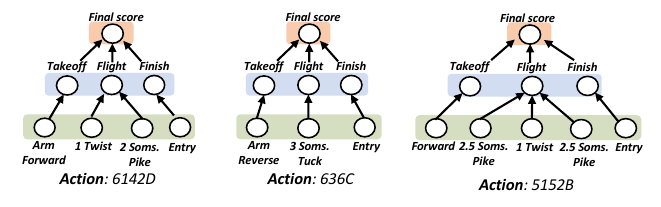}%
    \caption{Sample DAGs for three diving actions in Finediving.}\label{fig:supp_dags}%
\end{figure*}

A key component of \name is the integration of a scoring rubric in the form of a directed acyclic graph (DAG). As mentioned in Sec.\ 3, we assume that each video contains a set of steps, and each action step is independently scored. Subsequently, a rule-based rubric is employed to aggregate individual scores and calculate a final quality score, in which steps might be grouped into intermediate stages. Our DAG representation applies to a broad range of technical skill assessment tools. 

We use the FINA diving manual~\cite{fina_manual} as an example to further illustrate our representation. For an individual dive, a sequence of action items is predetermined, with various combinations leading to different difficulty degrees (see examples in Table~\ref{fig:diff-deg}). A panel of 7 to 11 judges will assess the dive, each assigning a score on a scale of 0-10. Judges will follow specified guidelines to evaluate the performance of the approach and takeoff (Table~\ref{fig:take-off}), the flight (Table~\ref{fig:flight}), and the entry into the water (Table~\ref{fig:entry}) to arrive at a final score. Other considerations \eg technique, execution and overall performance may be taken into account. The top two and the bottom two scores are discarded. The remaining scores are summed up and multiplied by the difficulty degree, yielding the final rating. Table~\ref{fig:score-sheet} illustrates a completed diving sheet from a real diving event. 

We note that \name does not employ the detailed exact rubric of FINA. Instead, our method follows its general structures by assuming key steps, the scoring of these steps, and the combination of individual scores --- a central concept in technical skill assessment~\cite{prassas2006biomechanical,waters1994applications,martin1997objective}.

\begin{table}[t]
    \centering
    \begin{minipage}{0.8\textwidth}
        \centering
        \caption{Guidelines for judging the entry into water~\cite{fina_manual}.}
        \includegraphics[width=\linewidth]{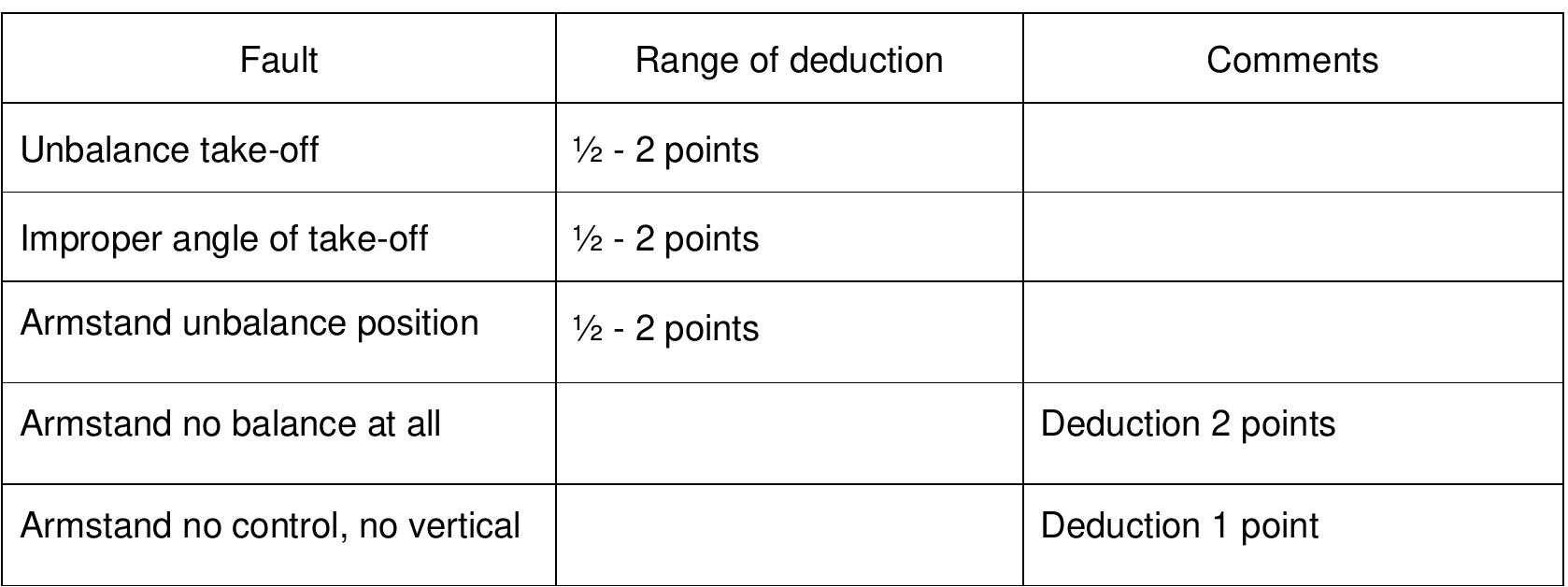}
        \label{fig:take-off}
    \end{minipage}\hfill
    \begin{minipage}{0.8\textwidth}
        \centering
        \caption{Guidelines for judging the approach and take-off~\cite{fina_manual}.}
        \includegraphics[width=\linewidth]{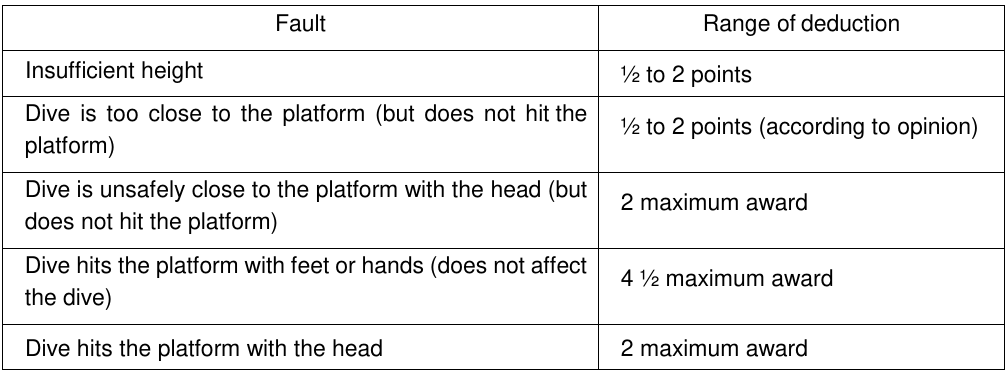}
        \label{fig:flight}
    \end{minipage}\hfill
    \begin{minipage}{0.8\textwidth}
        \centering
        \caption{Guidelines for judging the flight~\cite{fina_manual}.}
        \includegraphics[width=\linewidth]{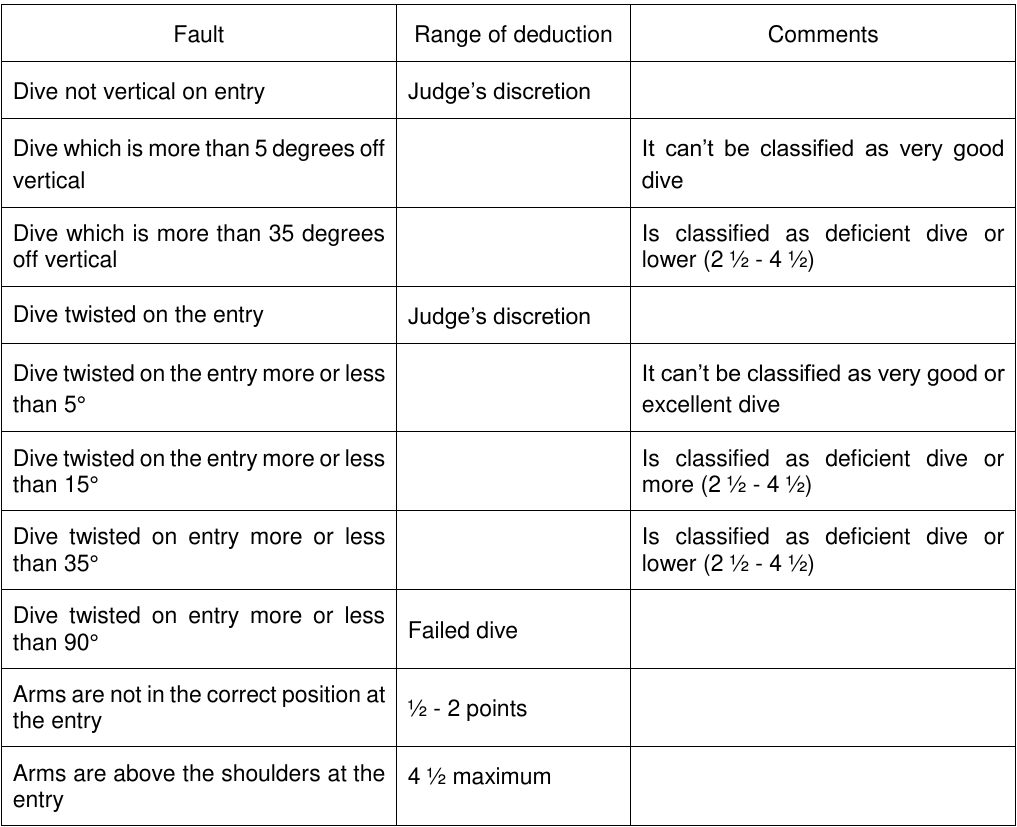}
        \label{fig:entry}
    \end{minipage}
\end{table}

\begin{table}[htbp]
    \centering
    \begin{minipage}{\textwidth}
        \centering
        \caption{How the difficulty degree for a dive is determined.}
        \includegraphics[width=\linewidth]{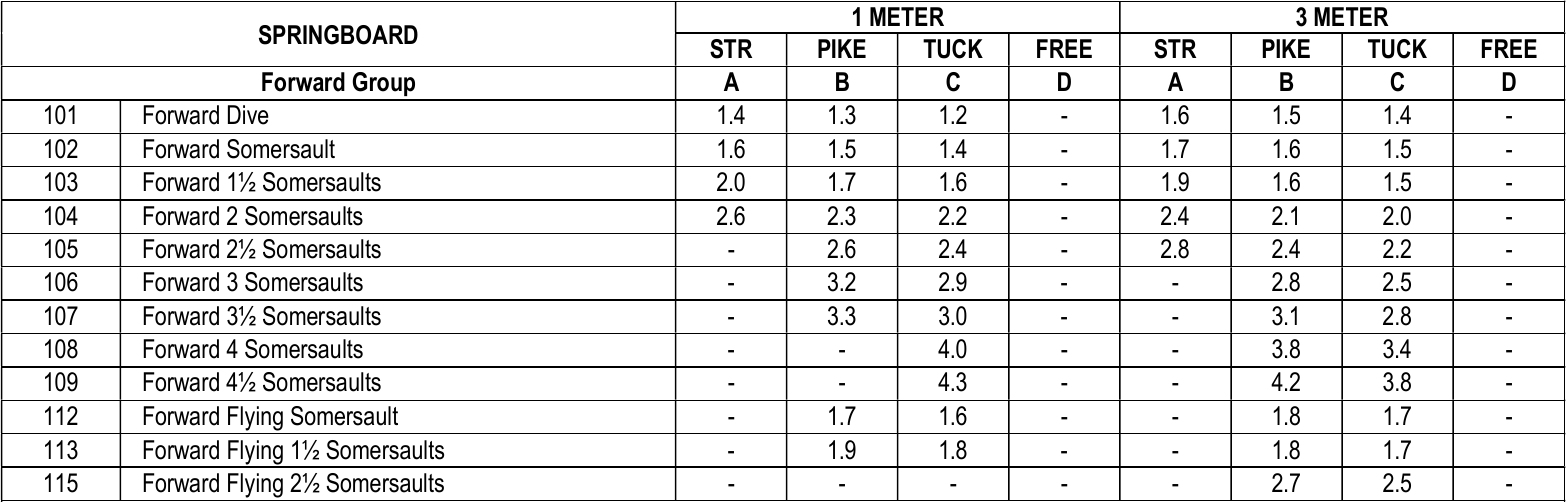}
        \label{fig:diff-deg}
    \end{minipage}\hfill
    \begin{minipage}{\textwidth}
        \centering
        \caption{Sample scoring sheet from a diving event}
        \includegraphics[width=\linewidth]{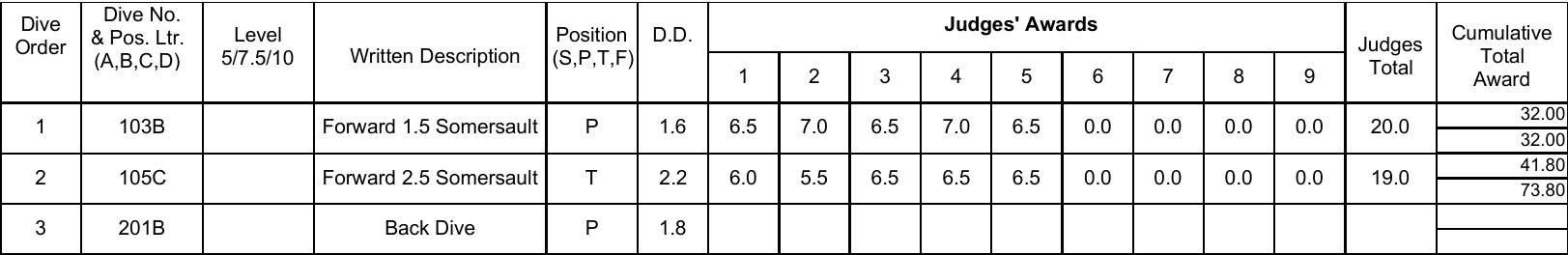}
        \label{fig:score-sheet}%
    \end{minipage}
\end{table}

\section{Further Discussions}\label{sec:discussion}

\subsection{Benchmark Settings}
We compare the setting of the baseline exemplar-based methods. Exemplar-based methods rely on exemplar videos and their corresponding scores during both the training and testing phases. Latest methods, such as TPT~\cite{bai2022action}, TSA~\cite{xu2022finediving}, and CoRE~\cite{xu_core_2021}, utilize dive numbers to select the exemplar videos and their associated quality scores. Notably, these dive numbers uniquely encode the presence and sequence of steps being executed, as shown in Table 7 of FineDiving~\cite{xu2022finediving}. Moreover, TSA incorporates not only the step presence but also precise start and end time stamps of each performed step. 

In sharp contrast, our method takes an input of step presence and ordering during training and inference, without the need for exemplar videos, their scores or timestamp data. Different from prior methods, our method additionally considers text descriptions of steps, obtained by intuitively expanding step names into simple sentences as shown in Tables~\ref{tab:mtl_aqa_textqueries},~\ref{tab:fine_aqa_textqueries}, and~\ref{tab:jigsaws_aqa_textqueries}. While our method assumes a pre-specified scoring rubric, this rubric is incorporated into our model design, and not as part of the input. 

\subsection{Trustworthy Visual Recognition}
Over the last decade, we have witnessed major advances in visual recognition with deep learning, leading to significantly improved results on public datasets. Despite the superior performance of these deep models, one remaining question is how users can trust their output, knowing that mistakes can be made by these models. This trustworthy visual recognition has multiple facets. We argue two of the key aspects are to provide credible confidence of the output and to consider a human-interpretable decision-making process. Our work in this paper takes a step towards these two critical aspects while addressing the challenging problem of video-based AQA. We presented a deep model for AQA that integrates the human score rubric and models the output confidence. Our method could facilitate high-stakes applications of AQA, including competitive sports and healthcare, where understanding and trusting the model's decisions are crucial, and low-confident samples can be readily passed to human experts.

\subsection{Ethical Concerns}
Our work focuses on the technical aspect of action quality assessment, and as such we do not anticipate major ethical concerns.

\begin{table}[]
\caption{Text descriptions for individual steps in MTL-AQA~\cite{parmar2019and}}
\centering
\resizebox{0.6\columnwidth}{!}
{
\footnotesize
\begin{tabular}{|l|p{0.6\linewidth}|}
\hline
\multicolumn{1}{|c|}{\textbf{Subaction}} &
  \multicolumn{1}{c|}{\textbf{Description}} \\ \hline
\multirow{4}{*}{Free} & \multicolumn{1}{p{0.6\linewidth}|}{In this position, athletes have the freedom to perform any combination of dives from various categories without any restrictions or limitations} \\ \hline
\multirow{4}{*}{Tuck} & \multicolumn{1}{p{0.6\linewidth}|}{In this position, athletes bring their knees to their chest and hold onto their shins while maintaining a compact shape throughout their dive}   \\ \hline
\multirow{4}{*}{Pike} & \multicolumn{1}{p{0.6\linewidth}|}{In this position, athletes maintain a straight body with their legs extended and their toes pointed out while bending at the waist to bring their hands toward their toes} \\ \hline
\multirow{4}{*}{Armstand} & \multicolumn{1}{p{0.6\linewidth}|}{In this position, athletes start by standing on their hands on the edge of the diving board and perform their dive while maintaining this handstand position} \\ \hline
\multirow{4}{*}{Inwards} & \multicolumn{1}{p{0.6\linewidth}|}{In this rotation type, athletes perform a forward-facing takeoff and rotate inward toward the diving board as they execute their dive}        \\ \hline
\multirow{4}{*}{Reverse} & \multicolumn{1}{p{0.6\linewidth}|}{In this rotation type, athletes perform a backward-facing takeoff and rotate backward away from the diving board as they execute their dive}  \\ \hline
\multirow{4}{*}{Backward} & \multicolumn{1}{p{0.6\linewidth}|}{In this rotation type, athletes perform a backward-facing takeoff and rotate backward toward the diving board as they execute their dive}    \\ \hline
\multirow{4}{*}{Forward} & \multicolumn{1}{p{0.6\linewidth}|}{In this rotation type, athletes perform a forward-facing takeoff and rotate forward away from the diving board as they execute their dive}    \\ \hline
\multirow{2}{*}{0.5 Somersault} & \multicolumn{1}{p{0.6\linewidth}|}{Athletes perform a half rotation in the air during their dive}                                                                         \\ \hline
\multirow{2}{*}{1 Somersault} & \multicolumn{1}{p{0.6\linewidth}|}{Athletes perform a full forward or backward rotation in the air during their dive}                                                       \\ \hline
\multirow{2}{*}{1.5 Somersault} & \multicolumn{1}{p{0.6\linewidth}|}{Athletes perform a full rotation and an additional half rotation in the air during their dive}                                         \\ \hline
\multirow{2}{*}{2 Somersault} & \multicolumn{1}{p{0.6\linewidth}|}{Athletes perform two full forward or backward rotations in the air during their dive}                                                    \\ \hline
\multirow{3}{*}{2.5 Somersault} & \multicolumn{1}{p{0.6\linewidth}|}{Athletes perform two full rotations and an additional half rotation in the air during their dive}                                      \\ \hline
\multirow{2}{*}{3 Somersault} & \multicolumn{1}{p{0.6\linewidth}|}{Athletes perform three full forward or backward rotations in the air during their dive}                                                  \\ \hline
\multirow{3}{*}{3.5 Somersault} & \multicolumn{1}{p{0.6\linewidth}|}{Athletes perform three full rotations and an additional half rotation in the air during their dive}                                    \\ \hline
\multirow{3}{*}{4.5 Somersault} & \multicolumn{1}{p{0.6\linewidth}|}{Athletes perform four full rotations and an additional half rotation in the air during their dive}                                     \\ \hline
\multirow{2}{*}{0.5 Twist} & \multicolumn{1}{p{0.6\linewidth}|}{Athletes perform a half twist in the air during their dive} \\ \hline
\multirow{2}{*}{1 Twist} & \multicolumn{1}{p{0.6\linewidth}|}{Athletes perform one full twist in the air during their dive} \\ \hline
\multirow{2}{*}{1.5 Twist} & \multicolumn{1}{p{0.6\linewidth}|}{Athletes perform one and a half twists in the air during their dive}                                                                        \\ \hline
\multirow{2}{*}{2 Twist} & \multicolumn{1}{p{0.6\linewidth}|}{Athletes perform two full twists in the air during their dive}                                                                                \\ \hline
\multirow{2}{*}{2.5 Twist} & \multicolumn{1}{p{0.6\linewidth}|}{Athletes perform two and a half twists in the air during their dive}                                                                        \\ \hline
\multirow{2}{*}{3 Twist} & \multicolumn{1}{p{0.6\linewidth}|}{Athletes perform three full twists in the air during their dive}                                                                              \\ \hline
\multirow{2}{*}{3.5 Twist} & \multicolumn{1}{p{0.6\linewidth}|}{Athletes perform three and a half twists in the air during their dive}                                                                      \\ \hline
\multirow{2}{*}{Entry} & \multicolumn{1}{p{0.6\linewidth}|}{A diving technique involving a entry into the water, typically performed at the end of a dive}                                                  \\ \hline
\end{tabular}
}
\label{tab:mtl_aqa_textqueries}
\end{table}

\begin{table}[]
\caption{Text descriptions for individual steps in FineDiving~\cite{xu2022finediving}}
\centering
\resizebox{0.6\columnwidth}{!}
{
\footnotesize
\begin{tabular}{|l|p{0.62\linewidth}|}
\hline
\multicolumn{1}{|c|}{\textbf{Subaction}} &
  \multicolumn{1}{c|}{\textbf{Description}} \\ \hline
\multirow{2}{*}{Forward} & \multicolumn{1}{p{0.62\linewidth}|}{A diving technique involving a front-facing takeoff and entry}                                                       \\ \hline
\multirow{2}{*}{Back} & \multicolumn{1}{p{0.62\linewidth}|}{A diving technique involving a back-facing takeoff and entry}                                                        \\ \hline
\multirow{2}{*}{Reverse} & \multicolumn{1}{p{0.62\linewidth}|}{A diving technique involving a back-facing takeoff and entry while rotating forward} \\ \hline
\multirow{2}{*}{Inward}               & \multicolumn{1}{p{0.62\linewidth}|}{A diving technique involving a front-facing takeoff and entry while rotating backwards}                              \\ \hline
\multirow{3}{*}{Arm Forward}          & \multicolumn{1}{p{0.62\linewidth}|}{A diving technique involving a front-facing takeoff and entry with arms extended and hands meeting above the head}   \\ \hline
\multirow{3}{*}{Arm Back}             & \multicolumn{1}{p{0.62\linewidth}|}{A diving technique involving a back-facing takeoff and entry with arms extended and hands meeting above the head}    \\ \hline
\multirow{3}{*}{Arm Reverse}          & \multicolumn{1}{p{0.62\linewidth}|}{A diving technique involving a back-facing takeoff and entry with arms extended and hands meeting above the head while rotating forward} \\ \hline
\multirow{3}{*}{1 Somersault Pike}    & \multicolumn{1}{p{0.62\linewidth}|}{A diving technique involving a takeoff and rotating forward to form a pike position with one somersault}             \\ \hline
\multirow{3}{*}{1.5 Somersaults Pike} & \multicolumn{1}{p{0.62\linewidth}|}{A diving technique involving a takeoff and rotating forward to form a pike position with one and a half somersaults} \\ \hline
\multirow{3}{*}{2 Somersaults Pike}   & \multicolumn{1}{p{0.62\linewidth}|}{A diving technique involving a takeoff and rotating forward to form a pike position with two somersaults}            \\ \hline
\multirow{3}{*}{2.5 Somersaults Pike} & \multicolumn{1}{p{0.62\linewidth}|}{A diving technique involving a takeoff and rotating forward to form a pike position with two and a half somersaults} \\ \hline
\multirow{3}{*}{3 Somersaults Pike}   & \multicolumn{1}{p{0.62\linewidth}|}{A diving technique involving a takeoff and rotating forward to form a pike position with three somersaults}          \\ \hline
\multirow{3}{*}{3.5 Somersaults Pike} & \multicolumn{1}{p{0.62\linewidth}|}{A diving technique involving a takeoff and rotating forward to form a pike position with three and a half somersaults}                   \\ \hline
\multirow{3}{*}{4.5 Somersaults Pike} & \multicolumn{1}{p{0.62\linewidth}|}{A diving technique involving a takeoff and rotating forward to form a pike position with four and a half somersaults}                    \\ \hline
\multirow{3}{*}{1.5 Somersaults Tuck} & \multicolumn{1}{p{0.62\linewidth}|}{A diving technique involving a takeoff and rotating forward to bend at the waist with one and a half somersaults}    \\ \hline
\multirow{3}{*}{2 Somersaults Tuck}   & \multicolumn{1}{p{0.62\linewidth}|}{A diving technique involving a takeoff and rotating forward to bend at the waist with two somersaults}               \\ \hline
\multirow{3}{*}{2.5 Somersaults Tuck} & \multicolumn{1}{p{0.62\linewidth}|}{A diving technique involving a takeoff and rotating forward to bend at the waist with two and a half somersaults}    \\ \hline
\multirow{3}{*}{3 Somersaults Tuck}   & \multicolumn{1}{p{0.62\linewidth}|}{A diving technique involving a takeoff and rotating forward to bend at the waist with three somersaults}             \\ \hline
\multirow{3}{*}{3.5 Somersaults Tuck} & \multicolumn{1}{p{0.62\linewidth}|}{A diving technique involving a takeoff and rotating forward to bend at the waist with three and a half somersaults}  \\ \hline
\multirow{3}{*}{4.5 Somersaults Tuck} & \multicolumn{1}{p{0.62\linewidth}|}{A diving technique involving a takeoff and rotating forward to bend at the waist with four and a half somersaults}   \\ \hline
\multirow{2}{*}{0.5 Twist}            & \multicolumn{1}{p{0.62\linewidth}|}{A diving technique involving a takeoff and half a twist before entering the water}                                   \\ \hline
\multirow{2}{*}{1 Twist}              & \multicolumn{1}{p{0.62\linewidth}|}{A diving technique involving a takeoff and one full twist before entering the water}                                 \\ \hline
\multirow{2}{*}{1.5 Twists}           & \multicolumn{1}{p{0.62\linewidth}|}{A diving technique involving a takeoff and one and a half twists before entering the water}                          \\ \hline
\multirow{2}{*}{2 Twists}             & \multicolumn{1}{p{0.62\linewidth}|}{A diving technique involving a takeoff and two full twists before entering the water}                                \\ \hline
\multirow{2}{*}{2.5 Twists}           & \multicolumn{1}{p{0.62\linewidth}|}{A diving technique involving a takeoff and two and a half twists before entering the water}                          \\ \hline
\multirow{2}{*}{3 Twists}             & \multicolumn{1}{p{0.62\linewidth}|}{A diving technique involving a takeoff with three twists before entering the water}                                  \\ \hline
\multirow{2}{*}{3.5 Twists}           & \multicolumn{1}{p{0.62\linewidth}|}{A diving technique involving a takeoff with three and a half twists before entering the water}                       \\ \hline
\multirow{2}{*}{Entry}                & \multicolumn{1}{p{0.62\linewidth}|}{A diving technique involving a entry into the water, typically performed at the end of a dive}                       \\ \hline
\multirow{3}{*}{0.5 Somersault Pike}  & \multicolumn{1}{p{0.62\linewidth}|}{A diving technique involving a take-off with half a somersault in the pike position before entering the water}       \\ \hline
\end{tabular}
}
\label{tab:fine_aqa_textqueries}
\end{table}

\begin{table*}[]
\caption{Text descriptions for individual steps in JIGSAWS~\cite{gao2014jhu}}
\centering
\resizebox{0.8\columnwidth}{!}
{
\footnotesize
\begin{tabular}{|l|l|p{0.7\linewidth}|}
\hline
\multicolumn{1}{|c|}{\textbf{Phase}} &
  \multicolumn{1}{c|}{\textbf{Subaction}} &
  \multicolumn{1}{c|}{\textbf{Description}} \\ \hline
\multirow{16}{*}{Suturing} &
  \multirow{1}{*}{G1} &
   \multicolumn{1}{p{0.7\linewidth}|}{Reaching for needle with right hand} \\ \cline{2-3}
 &
  \multirow{1}{*}{G2} &
   \multicolumn{1}{p{0.7\linewidth}|}{Positioning a needle to adjust its placement in a particular location or orientation} \\ \cline{2-3}
 &
  \multirow{2}{*}{G3} &
   \multicolumn{1}{p{0.7\linewidth}|}{Pushing a needle through tissue which involves applying force to the needle in order to penetrate and pass through bodily tissue} \\ \cline{2-3}
 &
  \multirow{1}{*}{G4} &
   \multicolumn{1}{p{0.7\linewidth}|}{Transfer a needle from the left hand to the right hand} \\ \cline{2-3}
 &
  \multirow{2}{*}{G5} &
   \multicolumn{1}{p{0.7\linewidth}|}{Moving to the center with the needle in grip which involves holding and manipulating the needle to direct it towards the central area of a target or site} \\ \cline{2-3}
 &
  \multirow{2}{*}{G6} &
   \multicolumn{1}{p{0.7\linewidth}|}{To pull a suture with the left hand is to use the left hand to apply tension and draw a length of suture thread through tissue} \\ \cline{2-3}
 &
  \multirow{1}{*}{G8} &
   \multicolumn{1}{p{0.7\linewidth}|}{Orienting a needle involves adjusting the position, angle, or direction of the needle} \\ \cline{2-3}
 &
  \multirow{2}{*}{G9} &
   \multicolumn{1}{p{0.7\linewidth}|}{The action of using the right hand to assist in tightening a suture which involves using the right hand to apply additional tension or pressure to the suture thread} \\ \cline{2-3}
 &
  \multirow{1}{*}{G10} &
   \multicolumn{1}{p{0.7\linewidth}|}{To loosen additional suture which involves manipulating the suture thread in order to reduce the tension or pressure that it is exerting on tissue} \\ \cline{2-3}
 &
  \multirow{2}{*}{G11} &
   \multicolumn{1}{p{0.7\linewidth}|}{Dropping the suture at the end and moving to the end points which involves releasing the suture thread from one hand and repositioning oneself or the needle to prepare for the next step in a medical procedure} \\ \hline
\multirow{10}{*}{Knot Tying} &
  \multirow{1}{*}{G1} &
   \multicolumn{1}{p{0.7\linewidth}|}{Reaching for needle with right hand} \\ \cline{2-3}
 &
  \multirow{2}{*}{G11} &
   \multicolumn{1}{p{0.7\linewidth}|}{Dropping the suture at the end and moving to the end points which involves releasing the suture thread from one hand and repositioning oneself or the needle to prepare for the next step in a medical procedure} \\ \cline{2-3}
 &
  \multirow{1}{*}{G12} &
   \multicolumn{1}{p{0.7\linewidth}|}{Reaching for a needle with the left hand involves extending the left arm and grasping the needle with the hand} \\ \cline{2-3}
 &
  \multirow{2}{*}{G13} &
   \multicolumn{1}{p{0.7\linewidth}|}{Making a C-loop around the right hand involves manipulating the suture thread in a circular motion to form a loop that encircles the fingers or hand of the right hand} \\ \cline{2-3}
 &
  \multirow{2}{*}{G14} &
   \multicolumn{1}{p{0.7\linewidth}|}{Reaching for a suture with the right hand involves extending the right arm and grasping the suture material with the hand} \\ \cline{2-3}
 &
  \multirow{2}{*}{G15} &
   \multicolumn{1}{p{0.7\linewidth}|}{Pulling a suture with both hands involves using both hands to apply tension and draw a length of suture thread through tissue} \\ \hline
\multirow{13}{*}{Needle Passing} &
  \multirow{1}{*}{G1} &
   \multicolumn{1}{p{0.7\linewidth}|}{Reaching for needle with right hand} \\ \cline{2-3}
 &
  \multirow{1}{*}{G2} &
   \multicolumn{1}{p{0.7\linewidth}|}{Positioning a needle to adjust its placement in a particular location or orientation} \\ \cline{2-3}
 &
  \multirow{2}{*}{G3} &
   \multicolumn{1}{p{0.7\linewidth}|}{Pushing a needle through tissue which involves applying force to the needle in order to penetrate and pass through bodily tissue} \\ \cline{2-3}
 &
  \multirow{2}{*}{G4} &
   \multicolumn{1}{p{0.7\linewidth}|}{Transfer a needle from the left hand to the right hand which involves moving the needle from one hand to the other} \\ \cline{2-3}
 &
  \multirow{2}{*}{G5} &
  \multicolumn{1}{p{0.7\linewidth}|}{Moving to the center with the needle in grip involves holding and manipulating the needle to direct it towards the central area of a target or site} \\ \cline{2-3}
 &
  \multirow{2}{*}{G6} &
   \multicolumn{1}{p{0.7\linewidth}|}{To pull a suture with the left hand is to use the left hand to apply tension and draw a length of suture thread through tissue} \\ \cline{2-3}
 &
  \multirow{1}{*}{G8} &
   \multicolumn{1}{p{0.7\linewidth}|}{Orienting a needle involves adjusting the position, angle, or direction of the needle} \\ \cline{2-3}
 &
  \multirow{2}{*}{G11} &
   \multicolumn{1}{p{0.7\linewidth}|}{Dropping the suture at the end and moving to the end points which involves releasing the suture thread from one hand and repositioning oneself or the needle to prepare for the next step in a medical procedure} \\ \hline
\end{tabular}
}
\label{tab:jigsaws_aqa_textqueries}
\end{table*}

\begin{table*}[]
\caption{Text descriptions for individual steps in Cataract~\cite{cataract101}}
\centering
\resizebox{0.8\columnwidth}{!}
{
\footnotesize
\begin{tabular}{|l|p{0.7\linewidth}|}
\hline
\multicolumn{1}{|c|}{\textbf{Subaction}} &
  \multicolumn{1}{c|}{\textbf{Description}} \\ \hline
\multirow{2}{*}{Incision}  &
  \multicolumn{1}{p{0.7\linewidth}|}{A sharp blade is used to create a precise cut through the cornea, which provides intraocular access for instruments. The paracentesis is followed by a clear cornea incision which is less than 3 mm wide and is large enough to insert the phaco handpiece}  \\ 
  
  \hline
\multirow{2}{*}{Viscous Agent Injection} & \multicolumn{1}{p{0.7\linewidth}|}{Viscous agent is injected to widen the anterior chamber and to protect the corneal endothelium and the intraocular structures. This is repeated before Phase 8 but is indistinguishable} \\ 
  \hline
  \multirow{3}{*}{Rhexis} & \multicolumn{1}{p{0.7\linewidth}|}{
  The anterior capsule of the lens is opened. The surgeon begins with a central radial cut. At the end of the cut, a tear is built and allows the anterior capsule to fold over itself. This tear is grasped and a flap is carried around in a circular way} \\
  \hline

  \multirow{2}{*}{Hydrodissection} &
  \multicolumn{1}{p{0.7\linewidth}|}{The surgeon injects electrolyte solution and epinephrin under the rhexis to separate the peripheral cortex of the lens from the capsule. This facilitates the rotation of the nucleus and hydrates the peripheral cortex} \\ \hline

  \multirow{3}{*}{Phacoemulsificiation} &
  \multicolumn{1}{p{0.7\linewidth}|}{With ultrasound power, the phaco tip emulsifies the anterior central cortex. A deep central linear groove through the nucleus is made and the lens is cracked into two parts. The lens is rotated and chopped into pieces, which can be emulsified. During this procedure, it is essential to keep the posterior capsule intact} \\ \hline

  \multirow{1}{*}{Irrigation and Aspiration} &
  \multicolumn{1}{p{0.7\linewidth}|}{Remaining parts of the cortex are extracted} \\ \hline
  
    \multirow{1}{*}{Capsule Polishing} & \multicolumn{1}{p{0.7\linewidth}|}{The posterior capsule is polished in order to avoid opacification of the capsule} \\ \hline

    \multirow{1}{*}{Lens Implant Setting-Up} &
  \multicolumn{1}{p{0.7\linewidth}|}{The folded artificial lens is inserted. The lens is slowly unfolding and is pushed into the capsular bag} \\ \hline

  \multirow{1}{*}{Viscous Agent Removal} &
  \multicolumn{1}{p{0.7\linewidth}|}{Viscous elastic agent is removed from anterior chamber and capsule bag} \\ \hline

  \multirow{2}{*}{Tonifying and Antibiotics} &
  \multicolumn{1}{p{0.7\linewidth}|}{The corneal incision is hydrated with electrolyte solution and antibiotics are injected. This induces temporary stromal swelling and closure of incision. Only if it leaks, a suture is required} \\ \hline
\end{tabular}
}
\label{tab:cataract_aqa_textqueries}
\end{table*}

\clearpage

{
    \bibliographystyle{splncs04}
    \bibliography{main}
}

\end{document}